\documentclass[acmsmall, preprint]{acmart}

\AtBeginDocument{%
  \providecommand\BibTeX{{%
    \normalfont B\kern-0.5em{\scshape i\kern-0.25em b}\kern-0.8em\TeX}}}

\setcopyright{acmcopyright}
\copyrightyear{2018}
\acmYear{2018}
\acmDOI{10.1145/1122445.1122456}

\acmConference[Woodstock '18]{Woodstock '18: ACM Symposium on Neural
  Gaze Detection}{June 03--05, 2018}{Woodstock, NY}
\acmBooktitle{Woodstock '18: ACM Symposium on Neural Gaze Detection,
  June 03--05, 2018, Woodstock, NY}
\acmPrice{15.00}
\acmISBN{978-1-4503-XXXX-X/18/06}

\usepackage{adjustbox}
\usepackage{multirow}
\usepackage{graphicx}
\usepackage{graphics}
\usepackage{array}
\usepackage{float}

\usepackage{algorithm}
\usepackage[noend]{algpseudocode}
\usepackage{amsmath}
\DeclareMathOperator*{\argmax}{argmax}
\begin{document}

\title{Evaluating Local Explanations using White-box Models}


\author{Amir Hossein Akhavan Rahnama}
\email{amiakh@kth.se}
\affiliation{%
  \institution{KTH Royal Institute of Technology}
  \city{Stockholm}
  \country{Sweden}
}

\author{Judith B\"utepage}
\email{judithb@kth.se}
\affiliation{%
  \institution{Spotify}
  \city{Stockholm}
  \country{Sweden}
}

\author{Pierre Geurts
}
\email{p.geurts@uliege.be}
\affiliation{%
  \institution{ University of Li\`ege}
  \city{ Li\`ege}
  \country{Belgium}
}

\author{Henrik B\"ostrom }
\email{bostromh@kth.se}
\affiliation{%
  \institution{KTH Royal Institute of Technology}
  \city{Stockholm}
  \country{Sweden}
}

\renewcommand{\shortauthors}{Akhavan Rahnama, et al.}




\begin{abstract}
    Evaluating explanation techniques using human subjects is costly, time-consuming and can lead to subjectivity in the assessments. To evaluate the accuracy of local explanations, we require access to the true feature importance scores for a given instance. However, the prediction function of a model usually does not decompose into linear additive terms that indicate how much a feature contributes to the output. In this work, we suggest to instead focus on the log odds ratio (LOR) of the prediction function, which naturally decomposes into additive terms for logistic regression and naive Bayes. We demonstrate how we can benchmark different explanation techniques in terms of their similarity to the LOR scores based on our proposed approach. In the experiments, we compare prominent local explanation techniques and find that the performance of the techniques can depend on the underlying model, the dataset, which data point is explained, the normalization of the data and the similarity metric. 
\end{abstract}



\maketitle
\section{Introduction}
\label{sec:intro}

% explanations are useful but we can not evaluate them
% global explanations
% local additive explanations
% reason for why you even need a surrogate is that the prediction function is not a linear additive term
% but for some simple models we can get it for the log odds
% how does the log odds relate to prediction function in binary case
% so 

% Definition 1 Additive feature attribution methods 

% problem statement: complicated, can not find additive contribution
% can we extract something from the prediction model that has it?
%. Yes

As machine learning models have become more complex, the need for techniques that explain the decision making process of the these \textit{black-boxes} has grown.  
To make the decision making process more accessible to humans, explanation techniques can be used to estimate the importance of input features to a model's output. %It is evident that explanations of machine learning models need to be accurate themselves. Explanation techniques that provide inaccurate explanations hinder the progress toward achieving accountability and transparency in machine learning \citep{rudin2018please}. 
Explanation techniques fall into two main categories: \textit{global} or \textit{local} explanations. Global explanations are a set of feature importance scores that are important for in the prediction of all instances in a dataset \citep{craven1995extracting}. Local explanations are the set of feature importance scores that are important in the prediction of a single instance from a dataset \citep{ribeiro2016model}. One prominent class of local explanation techniques are \textit{local additive} explanations that fit a linear - or additive - model to the predictions of the underlying model for a given instance \citep{lundberg2017unified}.  

\begin{figure}[t]
 \centering
  \includegraphics[scale=0.5]{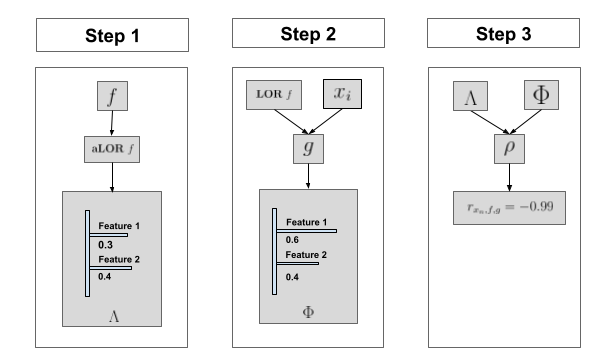}
 \caption{A flowchart of our proposed evaluation procedure is shown in this Figure. We begin by extracting the additive log odds ratio (aLOR) $\Lambda$ from model $f$ (Step 1). After that, we obtain explanation $\Phi$ using model-agnostic explanation $g$ for a single data point $x_i$  (Step 2)). Lastly, the accuracy $r_{x_i, f, g}$ of $g$ is computed as the Spearman's rank correlation  $\rho$ between $\Phi$ and $\Lambda$  (Step 3). % In this example, the explanations $\Phi$ are ranked in reversed order with respect to $\lambda$ and therefore the accuracy of the explanation is -0.99
 }
  \label{fig:procedure}
 \end{figure}
In order to evaluate the accuracy of explanations, two categories of evaluation methods have been proposed: human-grounded and functionally-grounded \citep{doshi2017towards}. \textit{Human-grounded} evaluation methods use the judgment of human subjects to evaluate the accuracy of explanations. The subjectivity of human judgment, time, and costs make human-grounded evaluations often infeasible. \textit{Functionally-grounded} methods use different proxy measures to evaluate explanation techniques systematically. Compared to human-grounded methods, functionally-grounded methods are fast and less costly. Functionally grounded evaluation methods are specifically useful in the early development of explanation techniques. 

As we discuss in Section~\ref{sec:background}, existing functionally-grounded methods cannot directly assess explanation accuracy. 
%One challenging problem of local explanation techniques is to evaluate their accuracy systematically \citep{hooker2018benchmark}. 
To assess accuracy directly, we need to have access to the ground truth \citep{bishop2006pattern}. In the context of explanations, this ground truth consists of the true feature importance scores for a given instance and prediction function. 
If we had access to a prediction function that decomposes into a linear weighted sum of the true importance scores, we would expect an accurate local additive explanation technique applied to this function to output scores that are not far from these true importance scores. The latter could thus be used to evaluate the accuracy of explanations. 
Since the prediction function of even simple classification models is not linear, we cannot directly assess accuracy based on the prediction function. 
In this work we propose to explain not the prediction function, but the log odds ratio (LOR) thereof. In particular, we focus on explaining two white-box models, logistic regression and naive Bayes, for which the log odds ratio decomposes into a linear sum of terms. These terms indicate how much a feature contributes to the prediction of one class versus another. We can thus interpret these additive log odds ratio (aLOR) terms as the true feature importance of the log odds ratio\footnote{The aLOR terms are also referred to as \textit{true importance scores} or \textit{aLOR scores} throughout our study.}. Given the true importance scores, we are now able to determine the accuracy of local additive explanation techniques. In our proposed method, we start by extracting the aLOR terms and local additive explanations based on the log odds ratio as shown in Step 1 and 2 in Figure~\ref{fig:procedure}. These can then be compared in terms of similarity as shown in Step 3 in Figure~\ref{fig:procedure}. We suggest to use our method as an initial verification of a new explanation technique which can be complemented with human- or other functionally-grounded evaluation techniques at later stages. 

We perform an extensive comparison of three local model-agnostic explanation techniques across 20 datasets; Local Interpretable Model-agnostic Explanations (LIME) \citep{ribeiro2016should}, SHapley Additive exPlanations (SHAP) \citep{lundberg2017unified} and Local Permutation Importance (LPI) \citep{casalicchio2018visualizing}. It is important to state that while LPI explanations are not additive, we are still interested to include LPI as a prominent local explanation technique.
Our key findings from this investigation are: 
% if too much space, then itemize
1) The performance of LIME and SHAP varies to a large extent across datasets, data points and underlying models. In some cases, the Spearman's rank correlation between LIME and SHAP explanations and aLOR terms is small or even negative. 2) LPI provides more accurate explanations relative to LIME and SHAP when the chosen similarity metric is Spearman's Rank Correlation. 3) LIME explanations provide more accurate explanations relative to SHAP and LPI when the chosen similarity metric is Euclidean or Cosine Similarity 4) The employed preprocessing technique has a large effect on the explanation accuracy, especially in the case of Logistic Regression. 5) The notions of robustness and accuracy of explanation techniques are complementary and orthogonal to one another

In the next section, we discuss related work on local model-agnostic explanation techniques and functionally-grounded evaluation. We then formally introduce the proposed evaluation method in Section \ref{sec:methodology}. In Section \ref{sec:experiments}, we empirically study the accuracy of explanation techniques in 20 datasets based on our proposed approach. We discuss the proposed method further in Section \ref{sec:discussion}, and finally, we summarize the main conclusions and point out directions for future research in Section \ref{sec:conclusion}.

\section{Local model-agnostic explanations and  functionally-grounded  evaluation}
\label{sec:background}

Local additive model-agnostic explanation techniques decompose the predicted probability of a black-box model into an additive sum of feature importance scores of a single instance \citep{lundberg2017unified}. Throughout this paper, the notion of local additive explanations and local additive model-agnostic explanations are used interchangeably. LIME and SHAP are examples of widely used techniques to obtain additive explanations. The explanations of LIME and SHAP consist of the weights of a local linear model that is trained to imitate the decision boundary of the black-box model for a given instance.   
In contrast, LPI defines the importance of a feature to be the average decrease in the prediction score when this feature value is randomly replaced by a different value \citep{casalicchio2018visualizing}. It is important to acknowledge that the feature importance scores of non-additive explanation techniques such as LPI have a different interpretation compared to additive explanation techniques.  

To evaluate the quality of explanations, functionally-grounded methods have been proposed. We have classified these method into two main categories: ablation and robustness evaluation techniques. \textit{Ablation techniques} measure the sensitivity of the model's predicted value after the removal of features with high importance scores \citep{hsieh2021evaluations}. The rationale behind this evaluation method is that the more substantial the change of the model's predicted value following the removal of important features, the more accurate are the explanations. Different ablation techniques are proposed in the literature of explainability which are different variations of the Occlusion Sensitivity method proposed in \citep{zeiler2014visualizing}. Ablation techniques therefore measure the accuracy of explanations indirectly, meaning without access to the ground truth.
One example of this technique is presented in \citep{hooker2018benchmark}. The authors propose RemOve And Retrain (ROAR) to evaluate different explanation techniques based on saliency maps. The black-box model is retrained on an auxiliary dataset that is striped of pixels deemed important by the explanations. The accuracy of the explanations is calculated as the difference between the predicted class probabilities of the original and the auxiliary instance. Methods such as ROAR are computationally expensive as the underlying model needs to be retrained on a large auxiliary data set. Another limitation of this evaluation method is that the new auxiliary data and retrained black-box model deviate from the original data and model and can exhibit new structures or properties. 

\textit{Robustness} of explanations on the other hand measures the difference in explanations of a single instance after adding noise or redundancy to the features of the explained instance \citep{alvarez2018robustness, camburu2019can}. The main rationale behind this category of evaluation techniques is that a robust explanation technique should provide explanations that are resilient with respect to the added noise and assign relatively low importance scores to redundant features.  
\citep{alvarez2018robustness} empirically show that some explanation techniques including LIME are not robust to Lipschitz noise. In \citep{camburu2019can}, the robustness of LIME, SHAP and Learn To eXplain (L2X) \citep{chen2018learning} is studied following the introduction of irrelevant features to the explained instance in a sentiment analysis task. According to the results, L2X explanations outperform other explanation techniques across different measures of robustness that are proposed in the study. One limitation of this category of evaluation techniques is that changing the feature values of the input instance can turn it into an out-of-distribution data point. This can lead to highly uncertain predictions by the black-box model which makes it impossible to distinguish between a failing black-box and a failing explanation \citep{ lakkaraju2020robust, rahnama2019study}. In section \ref{sec:experiment:robustness}, we perform experiments that measure the robustness of the explanation techniques in this study.

Both classes of techniques suffer from the fact that there is no objective way to directly derive the accuracy of an explanation solely based on the change of the black-box's output. Changing a feature value deemed important by an explanation does not necessarily lead to a change in the model's output. Similarly, if changing the feature moves the data point off the training data manifold, the output of the black-box might change dramatically \citep{lipton2018detecting}.

In contrast to ablation and robustness methods, we propose an evaluation method that extracts the true importance scores from the LOR and compares them directly to local explanations obtained from other explanation techniques. 
%In the next section, a summary of related work with a focus on the evaluation of different explanation techniques is presented. In Section \ref{sec:methodology}, we discuss how to extract model-intrinsic explanations from Logistic Regression and Naive Bayes. Section \ref{sec:experiments} contains  the evaluation and comparison of different model-agnostic techniques. In Section \ref{sec:conclusion}, we discuss our findings and include future directions for possible extensions of this work. 

 %\begin{figure}
 %\centering
 % \includegraphics[scale=0.25]{figs/lime_shap_exp.png}
 %\caption{LIME and SHAP Explanations of the Logistic Regression model trained on Pima Indians data set. The test instance explained is classified by the model to class \textit{Has Diabetes} with probability score of 0.64. LIME and SHAP (approximately) decompose the predicted value into a sum of importance scores for each feature in the data set.}
 % \label{fig:lime_shap_lpi_example}
 %\end{figure}
%\input{inputs/related}
\section{Methodology}
\label{sec:methodology}
\begin{figure}
 \centering
  \includegraphics[width=0.49\textwidth]{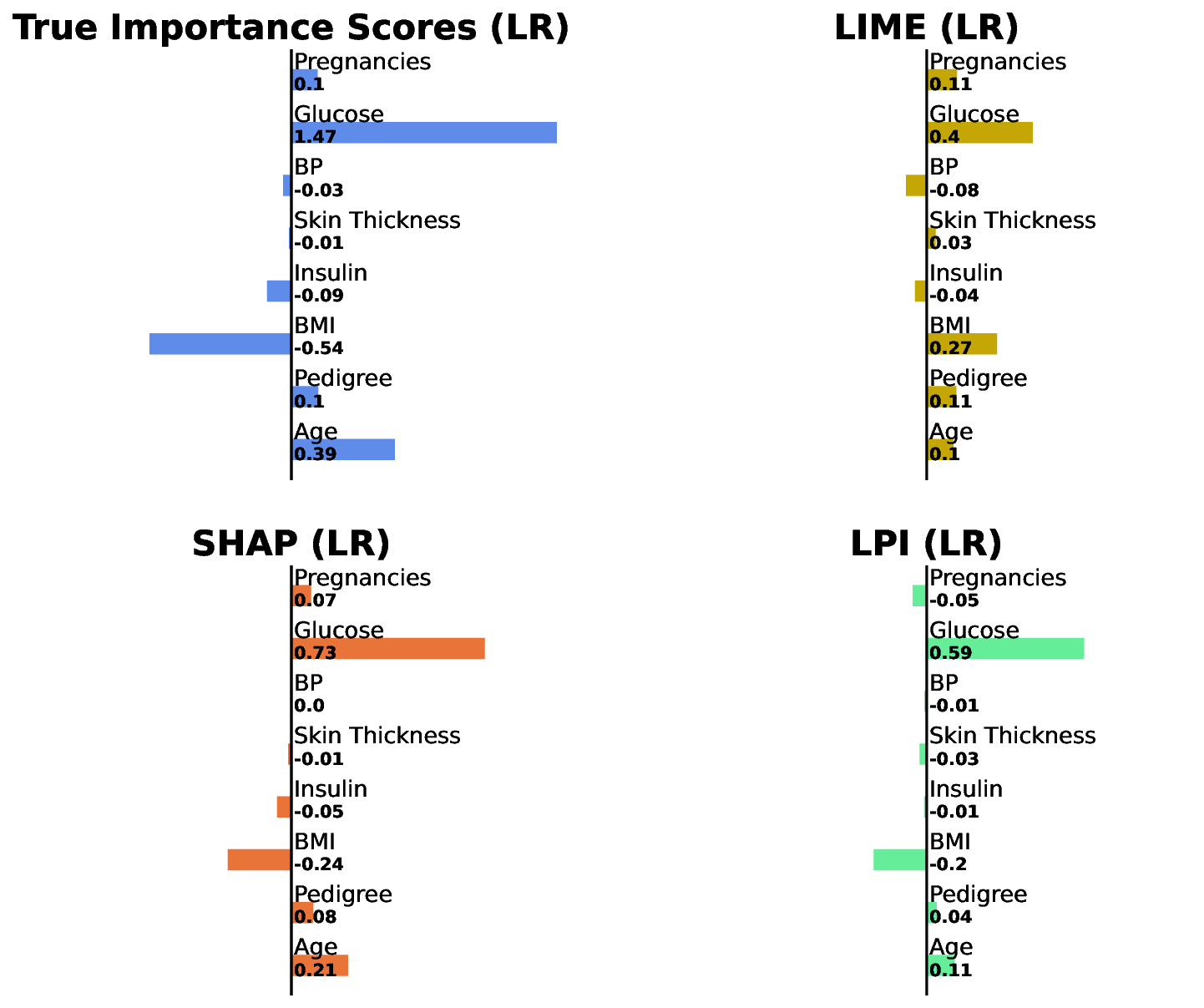}
 \caption{The feature importance scores of aLOR as well as LIME, SHAP and LPI explanations for a single instance from the Pima Indians data set when explaining a logistic regression prediction.}
  \label{fig:all_exps}
 \end{figure}

To evaluate the accuracy of a local model-agnostic explanation technique, we propose to extract the true importance scores of the LOR of two white-box models, logistic regression and naive Bayes. These scores are compared to local additive explanations obtained from the LOR of these white-box models (see Section \ref{sec:ground_truth}). We use Euclidean and Cosine Similarity along with Spearman's Rank correlation (see Section \ref{sec:evaluation}) to assess the accuracy of the explanations with respect to the true importance scores. If the explanations correlate highly with the true importance scores, we can conclude that the explanation is accurate. Figure \ref{fig:all_exps} shows a visualization of the true importance scores of a single instance as well as LIME, SHAP and LPI explanations.

\subsection{Local additive explanations}
\label{sec:notation}
Since LIME and SHAP are examples of local additive explanations, we present a formalization of local additive explanations based on the notation used in \citep{lundberg2017unified}. Let $X \in \mathbb{R}^{N \times M}$ be a matrix of $N$ data points with $M$ corresponding features and $Y \in \mathbb{R}^{N}$ be a vector of binary labels. Suppose $x_n \in \mathbb{R}^{M}$ is an instance of $X$ and let $x_n^m$ denote the $mth$ feature of $x_n$. 
 Assume that we want to explain the prediction of model $f$ for a given instance $x_n$.  The local additive explanation $g$ then has the form:
 \begin{equation}
 \label{eq:feature_additive}
 g(\hat{x}_n) = \phi^0  + \sum_{m=1}^M \phi^m\hat{x}_n^m,
 \end{equation}
 where $\hat{x}_n$ is a simplified version of $x_n$ and the prediction $g(\hat{x}_n)$ should be close to the prediction of the original function $f(x_n)$, i.e., $g(\hat{x}_n) \approx f(x_n)$. We denote the explanation by $\Phi_n = \{(x_n^m, \phi_n^m)|m=1, ..., M\}$.
 %In local additive explanations, the conditional probability $P $ of model $f$ decomposes the predicted class score for a single instance into a linear additive combination. Let $p_{n}^c$  be the predicted probability score for instance $x_n$ by model $f$ with respect to class $c \in (0,1)$, hence: $p_{n}^c =\phi_n^1 + ... + \phi_n^M$, where $\phi_n^m$ is the importance score for feature $m$.
 %Thus explanations can be represented as $\Phi_n = \{(x_m, \phi_m)|m=1, ..., M\}$. 

 In practice, a subset of $\Phi_n$ is selected including the top-$K$ ranked elements based on the absolute value of importance scores, where $K \ll M$ \citep{ribeiro2016should}. It is important to note that we extract the local model-agnostic explanations based on the log odds ratio of a single instance with respect to a specific class instead of the class probability scores $f(x_n)$. This is necessary as the extracted LOR scores are based on the log odds ratio.

\subsection{Local Permutation Importance}
\label{sec:lpr}

The core idea behind LPI \citep{casalicchio2018visualizing} is that the importance of a certain feature can be estimated by the average change of a black-box's prediction given that the value of this feature is replaced by another value. To change the feature value, LPI randomly permutates feature values of a single dimension across all data points. 

%to estimate the importance of a feature by measuring how much the black-box's prediction changes if the value of this feature changes. 

%calculate the importance of each feature as the average change in the model's predicted value after the feature value is replaced by its randomly permuted value.

%from the prediction of a model following a permutation on its corresponding feature values that are available in the dataset. 

%For this, the importance of each feature is calculated as the average change in model's predicted value after the feature value is replaced by its randomly permuted value. 

We calculate LPI as follows. Let $\pi$ be a random permutation of the index sequence $\langle 1, \ldots, N\rangle$, and let $\pi_i$ denote the position of index $i$ in $\pi$. The importance of feature $j$ at $x_n$ is then defined as:

%\begin{equation*}
%\gamma = 
%\begin{pmatrix}
%    1 & \cdots & N\\
%    i_1 & \cdots & i_N
%\end{pmatrix}
%\end{equation*}

\begin{equation}
    \Phi_n^j = \frac{1}{N} \sum_{k=1}^N(f({\hat{x}_k}) - f(x_n))
\end{equation}
where $\hat{x}_k$ is defined as follows:

\begin{equation}
    \hat{x}_{k}^l = 
    \begin{cases} 
      x_{n}^l & l \neq j \\
      x_{\pi_k}^j & l = j,
   \end{cases}
\end{equation}
where $k \in [1, N]$ and $l \in [1, M]$.
In simpler terms, $\hat{x}_{k}$ is equal to $x_n$ except that the value of the $j$th feature is replaced by $x_{\pi_k}^j$. It is noteworthy that we make use of the model's log odds ratio of class $c$ instead of the predicted values $f(x_n)$.

\subsection{True Importance Scores}
\label{sec:ground_truth}
The weights of logistic regression and naive Bayes models can be used as global explanations to understand which features are important in their learning and prediction process \citep{freitas2014comprehensible}. In our study, we propose to obtain the feature importance scores for a single data point from these models. Since the prediction functions of logistic regression and naive Bayes do not naturally decompose into additive terms, we turn the task of explaining the models' prediction function into the task of explaining the log odds ratio functions. We demonstrate that the log odds ratio can indeed decompose into additive parts for logistic regression and naive Bayes. As a result, each of the additive parts corresponds to a single feature and expresses how much the feature contributes to predicted log odds ratio of one class versus another. We call these terms {\it additive Log Odds Ratio} (aLOR) scores. Please note that while the notation in this section is based on a binary setting, however our proposed approach is extended to a multi-class settings using a one-vs-rest (OvR) method.

\subsubsection{Logistic regression}
\label{sec:method:lg}
Given weights $w \in \mathbb{R}^{M+1}$  and an instance $x_n \in \mathbb{R}^{M}$, the prediction function of a logistic regression model is defined as
	\begin{equation}
	\label{eq:pr_lr}
        P(y_n = c | x_n, w) = \frac{1}{1 + \textrm{e}^{{-\sum_{m=0}^{M} w^m x_n^m}}},
    \end{equation}
where we define $x_n^0 = 1$.
We can derive an additive decomposition of the prediction function by using the log odds ratio for $x_n$ with respect to class $c \in (0,1)$:

\begin{equation}
\label{eq:log_lr}
     \textrm{log}\frac{P(y_n = c|x_n, w )}{P(y_n = \neg c|x_n, w )} = \sum_{m=0}^{M} w^m x_n^m,
\end{equation}
where $\neg c$ is the complement of class $c$ and $\lambda_n^m = w^m x_n^m$ is the local importance score for feature $m$. The extracted  aLOR scores are denoted by $\Lambda_n = (\lambda_n^1, ..., \lambda_n^M)\in \mathbb{R}^{M}$. 
 
\subsubsection{Naive Bayes}
\label{sec:method:nbayes}
Given input $x_n$ and a mean and variance vector, $\mu_c \in \mathbb{R}^{M}$ and $\sigma_c  \in \mathbb{R}^{M}$,  the prediction function of  naive Bayes is defined as
\begin{equation}
    P(y_n = c | x_n) = \frac{P(x_n|y_n = c ) P(y_n = c)}{P(x_n)}, 
\end{equation} 
where the likelihood is given by
\begin{equation}
    P(x_n^m|y_n = c )  = \mathcal{N}(x_n^m|\mu_c^m, \sigma_c^m).
\end{equation}

Similar to the case of logistic regression, the prediction function does not naturally decompose into additive parts. However, the log odds ratio for an instance $x_n$ with respect to class $c$ has an intrinsic natural additive decomposition,

\begin{align}
\label{eq:log_naive}
	\textrm{log} \frac{P(y_n = c|x_n)}{P(y_n = \neg c|x_n)}  =  \sum_{m=1}^{M} \textrm{log} \frac{ \mathcal{N}(x_n^m|\mu_c^m, \sigma_c^m) }{\mathcal{N}(x_n^m|\mu_{\neg c}^m, \sigma_{\neg c}^m)} + const.
\end{align}
where $const = \textrm{log}\frac{P(y_n=c)}{P(y_n=\neg c)}$. Based on this, the importance score for feature $m$ is $\lambda_n^m = \textrm{log} \frac{ \mathcal{N}(x_n^m|\mu_c^m, \sigma_c^m) }{\mathcal{N}(x_n^m|\mu_{\neg c}^m, \sigma_{\neg c}^m)} $.

%%%%%%%%% TOREAD: %%%%%%%%%%%%%%%%%%
In our comparison, we compare each $\Lambda_j$ term with $\phi^j\hat{x}_n^j$ from Equation \ref{eq:feature_additive} since our proposed decomposition terms $\Lambda_j$ include two terms, namely the weight times the explained input. In each decomposed terms from Equation \ref{eq:feature_additive}, $\hat{x}$ is a simplified binary vector of one. This is due to the fact that local additive explanation do not provide an explanation for features with zero values \cite{lundberg2017unified,ribeiro2016should}. 
%%%%%%%%%%%%%%%%%%%%%%%%%%%%%%
 
\subsubsection{Example}
To make our idea more tangible, we train a logistic regression model with L2 regularization on a 2-dimensional discrete \textit{logical AND} function that returns one if both inputs are one and zero otherwise. The parameters of the logistic regression model are $w^1 = 0.422$ and $w^2 = 0.422$ with intercept value $w_0 = 0.69$. These parameters show that the model correctly learned that both features are equally important on a global level. For $x_0$ with $x_0^1 = 1$ and $x_0^2 = 0$ the model predicts $P(y_0=1|x_0) = 0.75$. Based on this, we can derive the log odds ratio for $x_n$ as
\begin{equation}
    \textrm{log}(\frac{0.75}{0.25}) = 0.69 +  0.422 \times 1 +  0.422 \times 0 \nonumber
\end{equation}
We can see that, whereas the first and second feature are equally important globally, the only feature that contributes to the log odds prediction is the first feature. The resulting aLOR scores consist of $\lambda_0^1  = 0.422$ and $\lambda_0^2 = 0$ the feature importance scores of the log odds ratio of $x_n$ with respect to class $1$. 
A similar example for naive Bayes can be found in the Appendix. 

% ---------- ---------- ---------- ---------- ---------- ---------- ---------- ---------- ---------- 
% ---------- ---------- ---------- ---------- ---------- ---------- ---------- ---------- ---------- 
% ---------- ---------- ---------- ---------- ---------- ---------- ---------- ---------- ---------- 
% ---------- ---------- ---------- ---------- ---------- ---------- ---------- ---------- ---------- 
\subsection{Evaluating explanation accuracy}
\label{sec:evaluation}

As discussed before, we suggest to use the aLOR scores as true importance scores when comparing explanation techniques. In order to measure the similarity between the true importance scores and different explanations, we use Euclidean and Cosine similarity along with Spearman's Rank correlation. We suggest to use the Spearman's Rank correlation as the primary measure of similarity (see Sections \ref{sec:similarity} and \ref{sec:example}). We explain how we can use the aLOR scores to evaluate the relative performance of explanation techniques in a systematic manner in Section \ref{sec:evaluationprocedure}. 
% ---------- ---------- ---------- ---------- ---------- ---------- ---------- ---------- ---------- 
% ---------- ---------- ---------- ---------- ---------- ---------- ---------- ---------- ---------- 
\subsubsection{Similarity between explanations}
\label{sec:similarity}

We measure accuracy in terms of how similar an explanation is to the true importance scores. Similarity between explanations can be measured in different ways. Several studies have used measures such as Cosine or Euclidean distance \citep{montavon2018methods, yang2019benchmarking} to measure the similarity of explanations. Similar to arguments in \cite{ghorbani2019interpretation}, we argue that the Spearman's Rank correlation is the most suitable measure for comparing explanations. This is because in numerous applications, explanations are presented as a small subset of ranked features based on their absolute importance scores \citep{ribeiro2016should}. Based on this, we may not need to pay attention to small changes in the importance scores that do not change the ranking of the features. In addition, a mistake in assigning low importance scores to a highly important features need to have a larger effect than of a assigning a wrong importance scores to modestly important features. The Spearman's Rank correlation coefficient is optimal in this setting as it is not affected by small, rank-preserving differences between explanations or differences in scale whereas such differences might affect cosine similarity or Euclidean distance \textit{drastically}. Cosine and Euclidean distances are useful in applications where the explanations are not ranked based on importance scores but represented as vectors that include all features. 

Additionally, the interpretation of explanations might differ across different types of explanation techniques. Rank correlation makes it possible to compare feature importance scores between additive and non-additive explanations techniques which otherwise cannot be compared directly. Lastly, correlation has a natural interpretation of how similar two vectors are and is confined to the interval between -1 and 1. In contrast to e.g.  Cosine and Euclidean similarity, correlation comes with interpretable measures of direction and strength. One drawback of using a rank-based measure is that it might be sensitive in case a dataset has many unimportant feature dimensions. In this case, the performance across all explanation techniques will be low as the ranking of unimportant features will vary randomly. As we show in the next section, at least one of the techniques usually has an average correlation of more than 0.4 for a given dataset, indicating that unimportant features do not impact explanation accuracy for the datasets chosen in this study. 

\subsubsection{Example}
\label{sec:example}

To understand the difference between the Euclidean and Cosine similarity along with  Spearman's rank correlation, imagine we want to compare two different explanations $\phi_1  = [0.21, 0.1, 0.32]$ and  $\phi_2 = [0.21, 0.3, 0.12]$ to the aLOR score $\lambda= [0.32, 0.2, 0.42]$. 

\noindent\begin{minipage}{.5\linewidth}
\begin{align}
    EuclideanS(\lambda, \phi_1) &= 0.179 \nonumber \\
    SpearmanC(\lambda, \phi_1) &= 1 \nonumber \\ \nonumber 
    CosineS(\lambda, \phi_1) &= 0.99 \nonumber \\ \nonumber 
\end{align}
\end{minipage}%
\begin{minipage}{.5\linewidth}
\begin{align}
    EuclideanS(\lambda, \phi_2) &= 0.28 \nonumber \\
    SpearmanC(\lambda, \phi_2) &= -1 \nonumber  \\
    CosineS(\lambda, \phi_2) &= 0.81\nonumber \\ \nonumber 
\end{align}
\end{minipage}%

As can be seen, the ranking of $\phi_1$ correlates perfectly with $\lambda$, while the ranking of $\phi_2$ is negatively correlated with $\lambda$. When using this rank-based metric, we can thus conclude that explanation $\phi_1$ is more accurate than $\phi_2$. 

%Not only Spearman's rank correlation correctly identifies that $\hat{\phi}$ is an inaccurate explanation, the relative difference between $\phi_1$, $\hat{\phi}_1$ is $2$ units whereas, the it is $0.1$ unit based on the Euclidean distance. 

%The Euclidean distance does not reflect the difference in the ranking and without any further information we can not draw any conclusion about which explanation is more accurate.  

% ---------- ---------- ---------- ---------- ---------- ---------- ---------- ---------- ---------- 
% Evaluation procedure
% ---------- ---------- ---------- ---------- ---------- ---------- ---------- ---------- ---------- 
\subsubsection{Evaluation procedure}
\label{sec:evaluationprocedure}

On a high level, to evaluate an explanation technique $g$ using the log odds ratio of white-box model $f$, we extract the aLOR scores $\Lambda$ and the local model-agnostic explanation $\Phi$ for a single instance $x_n$. We then compute the similarity between $\Lambda$ and $\Phi$ using  Spearman's rank correlation $r_{x_n, f,g} = \rho(|\Lambda|, |\Phi|)$. We measure the Spearman's rank correlation based on the absolute importance scores since that is how explanations are presented \citep{ribeiro2016should}. This procedure is detailed in Algorithm \ref{alg:approach}.
 
In general, we are interested in comparing different explanation techniques based on their similarity to the aLOR scores extracted from a white-box model $f$ across several datasets. %Performing experiments on numerous data sets can provide a more generalized perspective on the quality of explanations. his is due to the fact that we are explaining models that are trained on disparate combinations of features each of which with different empirical distributions. 
Let $d_i$ be the $i$th dataset of $T$ datasets. Suppose we obtain $R^{i}_{f, g} = \{r_{x_n,f, g}| n=1, ..., N_i\}$  by applying Algorithm \ref{alg:approach} to $N_i$ test instances of the $i$th  dataset for each explanation technique $g$. The average value of $R^{i}_{f, g}$ can be seen as a measure of how accurate the explanation technique $g$ is with respect to the aLOR scores. Needless to say, the higher the average values of $R_{f, g}$ are, the more accurate are the explanations $g$ when explaining model $f$. 

\begin{algorithm}[b!]
    \caption{Evaluating Explanations}
    \label{alg:approach}
    \begin{flushleft}
    \hspace*{2pt} \textbf{Input} $x_n$: instance\\
    \hspace*{2pt} \textbf{Functions} $f$: white-box model, $g$: explanation method, $t$: function to extract aLOR scores \\ 
    \hspace*{52pt}$\rho$: Spearman's rank correlation\\
    \hspace*{2pt} \textbf{Output} $r_{x_n,f, g}$: correlation value  
    \end{flushleft}
    \begin{algorithmic}[1]
    \State $\Phi  \gets g(f, x_n)$
    \State $\Lambda \gets t(f, x_n)$
    \State $r_{x_n, f, g} \gets \rho(|\Lambda|, |\Phi|)$
    \end{algorithmic}
\end{algorithm}

\subsection{Explanation Robustness}
\label{sec:robustness}
In this section, we present formal definitions of the robustness measures that are used in our experiments. 

\textit{Local Lipschitz} was proposed by \cite{alvarez2018robustness} to measure the maximum change in the predicted probability score of a black-box model in a neighbourhood around the explained instance. Let $X = \{x_i\}_{i=1}^{n}$ denote a sample of inputs and for every $x_i \in X$ let its neighbourhood be

\begin{equation}
    N(x_i) = \{x_j \in X | \quad \| x_i - x_j \| \leq \epsilon\}.
\end{equation}

$N(x_i)$ can also be created by adding Gaussian noise to the input sample $x_i$ and there is no need to set an $\epsilon$ in this case (see \cite{alvarez2018robustness} for details). Local Lipschitz $\widetilde{L}(x_i)$ is then defined as

\begin{equation}
\widetilde{L}(x_i) = \argmax_{x_j \in N(x_i)} \frac{\| f(x_i) - f(x_j )\| }{\| x_i - x_j\| }
\end{equation}
where $f$ is the black-box model that outputs a probability score for a designated class. While there is no ideal value, lower Local Lipschitz values indicate that an explanation technique is more robust. 

We consider two additional measures of robustness, namely $\textrm{Robustness}-S_r$ and $\textrm{Robustness}-\hat{S}_r$\footnote{These measures are refereed to as $\textrm{R}-S_r$ and $\textrm{R}-\bar{S}_r$ for brevity in Table \ref{table:final_robustness}} that are originally proposed by \cite{fong2017interpretable, samek2016evaluating}. Our notation is similar to that in \cite{hsieh2021evaluations}. Let $S_{r} \subset U$ be the set of important features selected by an explanation technique and let $\bar{S}_r = U - S_r$. $\textrm{Robustness}-S_r$ measures the change in the predicted probability scores of a black-box after the replacement of feature values in $S_r$ with a baseline value. Similarly, $\textrm{Robustness}-\bar{S}_r$ reflects the change in the predicted probability score of a black-box following the replacement of feature values in $\bar{S}_r$ with a baseline. The baseline value can be binary or the average value of the corresponding feature in the training or validation set. Similar to the case of Local Lipschitz, there are no ideal  optimal values for these robustness measures. It is then defined that robust explanations have relatively large $\textrm{Robustness}-S_r$ values and low $\textrm{Robustness}-\bar{S}_r$.

\begin{table*}[ht]

%\begin{minipage}[c]{0.67\textwidth}

\caption{Average Spearman's rank correlation between the true importance scores and local explanations for LIME, SHAP and LPI explanations when explaining Logistic Regression and na\"ive Bayes Models. Bold values indicate the explanation technique with the highest average correlation. }
\begin{tabular}{l|rrr|rrr}
    \toprule
     Model $\xrightarrow{}$&  \multicolumn{3}{c}{Logistic Regression} &  \multicolumn{3}{|c}{Na\"ive Bayes} \\ \hline
    Dataset   &   LIME &   SHAP & LPI &   LIME &   SHAP & LPI\\ \hline
     Adult              & 0.27  &  0.221 & 0.203 &  0.724 & 0.401 & 0.752 \\
     Attrition          & 0.502 &  0.194 & 0.423 &  0.261 & 0.297 & 0.297 \\
     Audit              & 0.515 &  0.511 & 0.703 &  0.194 & 0.382 & 0.55  \\
     Banking            & 0.226 &  0.477 & 0.338 &  0.486 & 0.088 & 0.848 \\
     Banknote           & 0.81  &  0.898 & 0.778 &  0.908 & 0.778 & 0.904 \\
     Breast Cancer      & 0.811 &  0.833 & 0.803 &  0.688 & 0.687 & 0.455 \\
     Churn              & 0.331 &  0.133 & 0.409 &  0.692 & 0.536 & 0.651 \\
     Donors             & 0.048 &  0.346 & 0.51  &  0.131 & 0.304 & 0.556 \\
     HR                 & 0.092 &  0.264 & 0.315 &  0.774 & 0.275 & 0.675 \\
     Haberman           & 0.74  &  0.377 & 0.708 &  0.786 & 0.026 & 0.877 \\
     Hattrick           & 0.403 &  0.657 & 0.637 &  0.586 & 0.564 & 0.446 \\
     Heart Disease      & 0.322 &  0.134 & 0.33  &  0.785 & 0.424 & 0.623 \\
     Insurance          & 0.595 & -0.301 & 0.604 &  0.616 & 0.199 & 0.644 \\
     Iris               & 0.728 &  0.792 & 0.848 &  0.848 & 0.892 & 0.78  \\
     Loan               & 0.398 &  0.33  & 0.453 &  0.311 & 0.463 & 0.463 \\
     Pima Indians       & 0.775 &  0.856 & 0.593 &  0.743 & 0.469 & 0.606 \\
     Seismic            & 0.501 &  0.511 & 0.497 &  0.773 & 0.198 & 0.716 \\
     Spambase           & 0.606 &  0.244 & 0.552 & -0.259 & 0.182 & 0.463 \\
     Thera              & 0.488 &  0.208 & 0.557 &  0.306 & 0.48  & 0.48  \\
     Titanic            & 0.779 &  0.245 & 0.74  &  0.695 & 0.366 & 0.794 \\
    \hline
      \midrule
    Average            & 0.497 &  0.396 & 0.55  &  0.552 & 0.401 & 0.629 \\
     Standard Deviation & 0.23  &  0.293 & 0.176 &  0.293 & 0.216 & 0.162 \\
 \bottomrule
\end{tabular}
 
  %\end{minipage}\hfill
  %\begin{minipage}[c]{0.2\textwidth}
  %  \caption{Average Spearman's rank correlation between the true importance scores and local explanations for LIME, SHAP and LPI explanations when explaining Logistic Regression and na\"ive Bayes Models. Bold values indicate the explanation technique with the highest average correlation. }
   \label{table:lime_shap_lpi}
 % \end{minipage}

\end{table*}

\section{Empirical Investigation}
\label{sec:experiments}

%In this section, we will present results from the empirical investigation. 
We start out by describing the experimental setup,
which is followed by results from comparing the selected explanations techniques against true importance scores. %As stated in Section \ref{sec:ground_truth}, we compare the log odds ratio of a single instance with respect to a specific class instead of the class probability scores. 
Finally, we investigate the effect of the choice of preprocessing technique, used prior to model generation.
%In \ref{sec:experiment:preprocessing}, we show that the comparison is substantially affected by the choice of which preprocessing technique is applied to the numerical features in each dataset. 

\subsection{Experimental setup}
\label{sec:experiment:setup}
In this section, we provide the results of our empirical investigation where we compare explanations of LIME, SHAP, and LPI when explaining logistic regression and naive Bayes models in terms of their accuracy\footnote{The code for experiments is available at: \url{https://anonymous.4open.science/r/Evaluation-Local-Explanations-Whitebox-Models-BFC5/}}.

We evaluate the proposed procedure on 20 different tabular datasets, all concerning classification tasks (binary and multi-class).  Unless otherwise stated, the numerical features are standardized and categorical features are encoded as one-hot vectors. For datasets for which no separate test set has been provided at the source, a random hold-out set of 25\% was used. In order to tune the hyper-parameters of the logistic regression models we use grid-search\footnote{Hyper parameters were chosen after 10 trials with the hyper-parameter space consisting of L1 and L2 regularization with the regularization parameter selected from a uniform distribution of values between 0 to 4.} . The Appendix includes the test accuracy of the logistic regression and naive Bayes models along with references to the datasets used in the experiments, which are all publicly available. 

For LIME and SHAP, we used the official Python packages TabularLIME \citep{ribeiro2016model} and KernelShap \citep{lundberg2017unified}. We would like to emphasize that KernelShap explainer is model-agnostic and it outputs approximated SHAP values. This is contrast to model-based explainers such as LinearSHAP where the SHAP values are analytically deducible from closed-form equations (see \citep{lundberg2017unified} for details). In our study, we are comparing model-agnostic explanations where the explainers have no assumptions on the class of machine learning models they are explaining. The number of samples generated for LIME and SHAP is 5000 as suggested in \cite{lundberg2017unified, ribeiro2016model}. In the case of LPI, it is the size of the training set as suggested in \citep{casalicchio2018visualizing}. This means the sample size of LPI is half the size of LIME and SHAP on average. As stated earlier, the feature importance scores are extracted from the log odds ratios of instances with respect to a specific class. Therefore we need to obtain the local explanations based on the log odds ratio of a given class. In the case of binary classification, we explain the class designated as one in the dataset to be in line with equations in Section \ref{sec:method:lg} and \ref{sec:method:nbayes}. In the case of multi-class classification, the predicted class by the underlying model is explained. The only modification we have made is to pass the log odds ratio function as a replacement of the prediction function when obtaining explanations. This is possible as in both packages one can pass in any desired prediction function when obtaining explanations. In the case of LPI, we have replicated the algorithm proposed in \citep{casalicchio2018visualizing} such that the importance scores are calculated based on the log odds ratio scores instead of the prediction function (see Section \ref{sec:lpr}).
 
\subsection{General performance}
\label{sec:experiment:logodds}
The average rank correlations for all datasets and explanation techniques are presented in Table \ref{table:lime_shap_lpi}. The first three columns list the results for the logistic regression white-box models and the last three columns list the equivalent for the naive Bayes white-box models. We also summarize the average rank correlation values across all datasets as well as the standard deviation in the last two rows. 

%\begin{figure}[b]
% \centering
%  \includegraphics[width=0.9\textwidth]{figs/odds_standard.png}
% \caption{Box-plots of Spearman's rank correlation of explanation techniques when the underlying model is Logistic Regression (top) and naive Bayes (bottom). The shaded area visualizes the threshold where $\textrm{corr} > 0.7$.}
%  \label{fig:tabular_odds}
% \end{figure}

Let us first focus on the explanations of the two local additive techniques, LIME and SHAP, obtained for the logistic regression models. LIME provides explanations with relatively low average rank correlation for several datasets, such as Adult, Churn, Donors and Heart Disease. On the other hand, the average rank correlation of SHAP explanations is substantially low for the Attrition, Donors, Insurance and HR datasets to name a few. The average rank correlation of LIME is larger than the average rank correlation of SHAP across all datasets. This suggests that LIME explanations are more accurate with respect to the aLOR scores for Logistic Regression in most cases. %However, there are numerous datasets that LIME explanations provide more accurate explanations than SHAP when explaining Logistic Regression, e.g. Adult, Attrition, Audit, Churn and etc.

Looking at the performance of LIME and SHAP in the context of the naive Bayes models, we can see that the performance of LIME and SHAP are both slightly improving compared to when using logistic regression. This suggests that the performance of the two explanation techniques might depend on the class of black-box (here white-box) models used. As a result, we can conclude that while model-agnostic explanations can provide explanations for all classes of machine learning models, they are more accurate for some classes of machine learning models. 

%The performance of LIME tend to vary to a higher degree over models and datasets compared to SHAP. For example, the average rank correlation of LIME for the Titanic dataset increases from $0.595$  when using logistic regression to $0.830$ when using naive Bayes respectively, while it is $0.797$ and $0.215$ for the Thera dataset.For example, the average rank correlation of LIME for the Titanic dataset is $0.595$ and $0.830$ for Logistic Regression and naive Bayes respectively, while it is $0.797$ and $0.215$ for the Thera dataset.

Turning our attention to LPI, the explanations of LPI result in higher average rank correlation across all datasets. This is quite surprising, as one would expect that additive explanations, as produced by LIME and SHAP, would outperform explanations that do not decompose into additive parts, such as the ones produced by LPI. We do not (yet) have a good explanation for this finding.

%\subsubsection{Variance across  and within  datasets}
%\label{sec:stdandhighcorrelation}

\begin{figure}[t]
 \centering
  \includegraphics[scale=0.3]{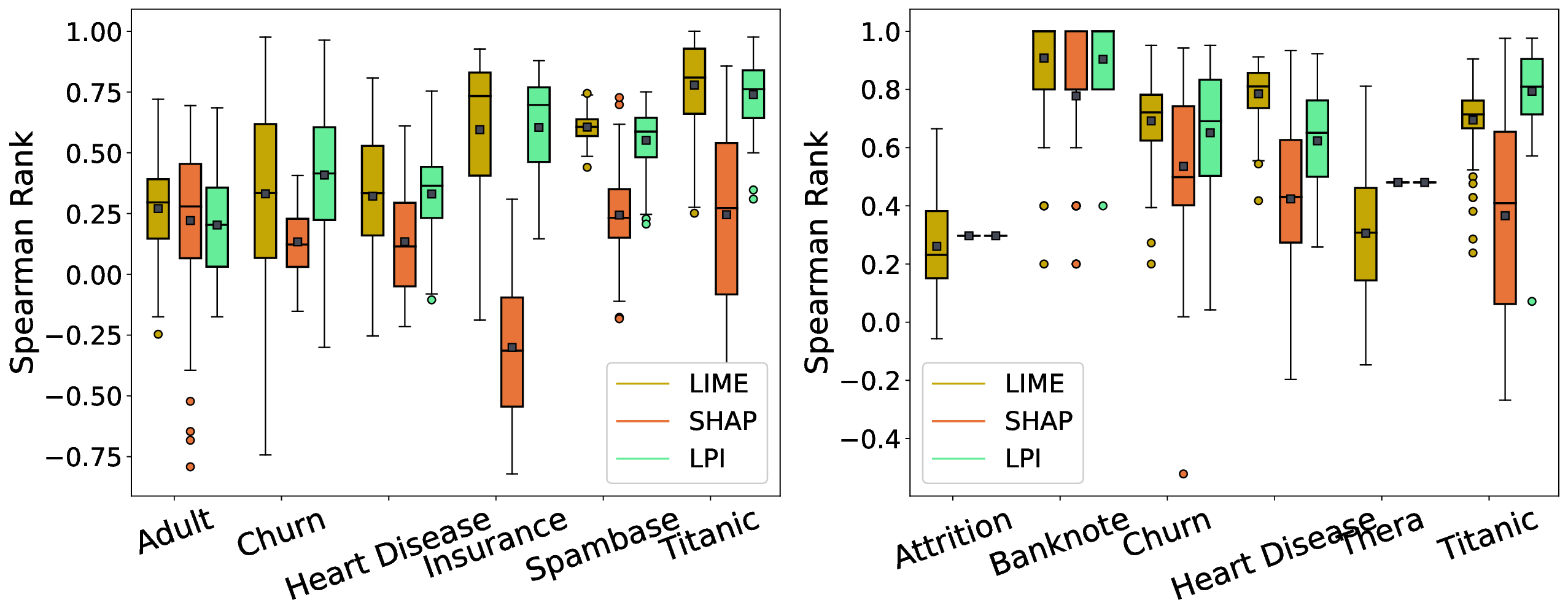}
 \caption{Box-plots of Spearman's rank correlation of explanation techniques when the underlying model is Logistic Regression (left) and naive Bayes (right) from a subset of datasets used.  The dark rectangles are indicators of average values in each box-plot. }
  \label{fig:tabular_boxplot}
 \end{figure}

That the performance of the explanation techniques is not constant across different datasets becomes apparent when we observe that the Spearman's rank correlation can vary to a high degree depending on the data point  within each dataset (Figure \ref{fig:tabular_boxplot}). For several datasets, the standard deviation of the correlation is substantially higher for LIME and SHAP compared to LPI. This means that the explanation techniques to some degree, can produce both highly correlated explanations for some instances of a dataset and weakly correlated or uncorrelated explanations for other instances using exactly the same underlying model. One possible implication of this phenomenon is that local explanation techniques are not able to explain all instances within a dataset accurately. We believe that the local explanation techniques need to develop criteria for cases in which the explanation technique cannot provide an explanation. Based on our knowledge, none of the explanation techniques in this study has such criteria. 

Table \ref{table:final_rank_norm} includes the results of similar experiments with different similarity measures. In the case of Logistic Regression, LIME explanations are most similar to LOR scores when using Cosine and Euclidean similarity. Similar conclusions can be made in the case of naive Bayes. The decrease in the accuracy of LPI explanations can be explained by the fact that, in contrast to LIME and SHAP, LPI's feature importance scores  are non-additive and can therefore not be compared to additive explanations directly in Euclidean space (see Section \ref{sec:lpr} and \ref{sec:similarity}). The result shows the strong contribution of the choice of a similarity metric on explanation accuracy. The choice of an ideal similarity measure is not an easy task as there is no unified agreement on the optimal measure for explanation accuracy and it is highly dependent on the application scenario \cite{ghorbani2019interpretation}. 

 \begin{table*}[th]
   \caption{The average of similarity between LOR scores and explanations across all data sets based on the different measures of similarity.}
  %\begin{minipage}[c]{0.6\textwidth}
  \centering
    \begin{tabular}{lrrr|rrr}
    \toprule
            & \multicolumn{3}{c}{Logistic Regression}  & \multicolumn{3}{c}{naive Bayes} \\
    \toprule
               &   LIME &   SHAP &   LPI & LIME &   SHAP &   LPI  \\
        \midrule
        Spearman's Correlation & 0.497 & 0.396 & \textbf{0.55}  & 0.552 & 0.401 & \textbf{0.629}  \\
        Euclidean Similarity       & \textbf{0.702} & 0.598 & 0.646 & \textbf{0.741} & 0.469 & 0.692  \\
        Cosine  Similarity     & \textbf{0.397} & 0.385 & 0.394 & \textbf{0.278} & 0.253 & 0.274    \\
    \bottomrule
    \end{tabular}
   % \end{minipage}\hfill
  %\begin{minipage}[c]{0.2\textwidth}
   
  \label{table:final_rank_norm}
  %\end{minipage}
\end{table*}

\subsection{Explanation Robustness}
\label{sec:experiment:robustness}
In this section, we share the results of empirical experiments on the robustness of LIME, SHAP and LPI explanations when explaining Logistic Regression and naive Bayes on our datasets. Local Lipschitz and $\textrm{Robustness}-\hat{S}_r$ and $\textrm{Robustness}-S_r$ are measures of robustness that are described (see Section \ref{sec:robustness}). To recapitulate, a robust explanation technique have relatively low values for Local Lipschitz and $\textrm{Robustness}-\hat{S}_r$ and relatively high values for $\textrm{Robustness}-S_r$. 

In Table \ref{table:final_robustness}, the average robustness values are shown across all data sets\footnote{The detailed results are reported in Section \ref{sec:appendix:robust} in the appendix}. $\textrm{Robustness}-S_r$, $\textrm{Robustness}-\hat{S}_r$ are based on the top-5 important features in the dataset. Local Lipschitz values are based on adding noise from a standard normal distribution with $|N(X)| = 3$.  LPI provides explanations that are more robust in terms of $\textrm{Robustness}-S_r$ and $\textrm{Robustness}-\bar{S}_r$ for both, Logistic Regression and naive Bayes models. For both explained models, SHAP explanations are the most robust in terms of Local Lipschitz measure. We see a different pattern here compared to Table \ref{table:final_rank_norm}. LIME appears to be the least robust explanation technique while simultaneously providing the most accurate explanations in terms of Euclidean and Cosine Similarity. LPI on the other hand seems to be the robust and accurate in terms of Spearman's Rank Correlation.

 \begin{table*}[th]
   \caption{The average robustness of explanations across all data sets .}
  \centering
    \begin{tabular}{lrrr|rrr}
    \toprule
            & \multicolumn{3}{c}{Logistic Regression}  & \multicolumn{3}{c}{naive Bayes} \\
    \toprule
               &   LIME &   SHAP &   LPI & LIME &   SHAP &   LPI  \\
        \midrule
         $\textrm{R}-S_r$ & 0.108 & 0.086 & \textbf{0.122} & 0.096 & 0.117 & \textbf{0.257} \\
         $\textrm{R}-\bar{S}_r$ & 0.038 & 0.049 & \textbf{0.033} & 0.084 & 0.082 & \textbf{0.061}    \\
         $\widetilde{L}$ & 2.17  & \textbf{0.172} & 0.191 & 2.621 & \textbf{0.441} & 0.896   \\
    \bottomrule
    \end{tabular}
   
  \label{table:final_robustness}
\end{table*}

\subsection{Preprocessing effects}
\label{sec:experiment:preprocessing}

% we have shown that correlation depends on dataset, white box model, data point, so what else does it depend on? 

% something that people often do, but sometimes do not even mention, is preprocessing

% prepossing is ralely looked at in the context of explanations / as a contributing factor of feature importance

% this can change the emperical feature distribution

% 
As described in Section \ref{sec:experiment:setup}, we standardized the numerical features in each dataset before training the logistic regression and naive Bayes models. In this section, we show that the choice of employed preprocessing technique  has a substantial effect on the accuracy of the generated explanations. We compare the performances from using three different preprocessing techniques: standardization, robust scaling and minmax (see \ref{appendix:prproc_def} for definition).

The first three columns in Table \ref{table:final_rank_preprocessing} show the average accuracy across all datasets when explaining logistic regression predictions. LIME and SHAP explanations provide the most accurate explanations in many cases when explaining Logistic Regression. The relative performance of explanation techniques can be highly dependent on the preprocessing method used regardless of the measures of similarity used. For example, the difference between performing Minmax or Standard preprocessing can have a substantial and decisive effect in choosing LIME over SHAP in cases where Euclidean or Cosine similarity is used as the similarity metric. 

The last three columns of the Table \ref{table:final_rank_preprocessing} show results for when explaining predictions of naive Bayes models. In this case, the choice of preprocessing technique has less of an effect on the relative performance of the three techniques. LIME and LPI outperform SHAP across all cases. %Based on our results, we suggest that practitioners choose the preprocessing technique that helps increase the accuracy of explanations when using a single explanations technique. 

Since the choice of preprocessing technique has only a very minor effect on the predictive performance of the underlying models across the datasets (see Table \ref{table:accuracy} in appendix), performance shifts of these models cannot explain the observed differences in explanation accuracy following the choice of preprocessing technique. One possible explanation is that explanation techniques have implicit assumptions on the empirical distributions of features.

 \begin{table*}[th]
   \caption{The average similarity across all datasets for each preprocessing techniques}
  \centering
    \begin{tabular}{lrrrrr|rrr}
            & &  & \multicolumn{3}{c}{Logistic Regression}  & \multicolumn{3}{|c}{naive Bayes} \\
        \toprule
            & &  Preprocessing &  LIME &   SHAP &   LPI & LIME &   SHAP &   LPI  \\
        \midrule
          \multirow{9}{*}{\rotatebox[origin=c]{90}{Similarity}} & \multirow{3}{*}{Spearman} & standard & 0.497 & 0.396 & \textbf{0.55}  & 0.552 & 0.401 & \textbf{0.629} \\
         & & minmax   & 0.15  & \textbf{0.397} & 0.165 & \textbf{0.641} & 0.414 & 0.626  \\ 
         & & robust   & 0.315 & \textbf{0.618} & 0.303 & 0.596 & 0.401 & \textbf{0.628} \\
         \cline{3-9}
         & \multirow{3}{*}{Cosine} & Standard & \textbf{0.702} & 0.598 & 0.646 & \textbf{0.741} & 0.469 & 0.692 \\
         & & Minmax   & 0.276 & \textbf{0.348} & 0.17  & \textbf{0.82}  & 0.478 & 0.691 \\
         & & Robust   & 0.52  & \textbf{0.72}  & 0.366 & \textbf{0.796} & 0.473 & 0.689 \\
        \cline{3-9}
         & \multirow{3}{*}{Euclidean} & Standard & \textbf{0.397} & 0.385 & 0.394 & \textbf{0.278} & 0.253 & 0.274 \\
         & & Minmax   & 0.325 & \textbf{0.387} & 0.336 & \textbf{0.282} & 0.251 & 0.273  \\
         & & Robust   & \textbf{0.398} & 0.434 & 0.376 & \textbf{0.277} & 0.253 & 0.274\\
    \bottomrule
    \end{tabular}
   
  \label{table:final_rank_preprocessing}
\end{table*}

\section{Discussion}
\label{sec:discussion}

Claiming to have derived the true feature importance for a given instance and underlying model, might appear to be unfounded. One could argue that there exist an infinite amount of linear combinations of features that add up to the predicted score, so why would the one presented in this paper be the \textit{true} decomposition. The beauty of explaining the log odds ratio is that we can extract the exact linear combination based on the model's parameters for a given instance, i.e., we can extract the \textit{only} decomposition that the model uses to generate the log odds ratio score.  

Another question that might occur to the reader is whether using simple models can tell us anything about the performance of more complex models. It is clear that good performance on simple tasks does not necessarily transfer to good performance on complex tasks. This means that an explanation technique, that has high accuracy according to our method, might not perform well on more complex tasks. The important aspect of using simple tasks however is to have access to a sanity check; \textit{if the explanation technique does not work on a simple task, then it is very unlikely that it will have high accuracy on complex tasks}. This means that functionally-grounded evaluation methods, such as the one presented here, can be seen as means to make the development of new candidate techniques more efficient; they allow for rejecting some candidate techniques early on. Promising candidates will most likely still need to be qualitatively evaluated in a user-centered context at later stages.     

\section{Concluding remarks}
\label{sec:conclusion}

In this study, we propose an evaluation procedure to directly measure the accuracy of local explanation when explaining the log odds ratio (LOR) scores of logistic regression and naive Bayes models. We showed that in contrast to prediction functions, the log odds ratio functions of these models have  intrinsic additive structures. We presented a comparison of explanation techniques based on the extracted true importance scores using Spearman's rank correlation, Euclidean and Cosine similarity. The results from the extensive empirical investigation showed that the quality of the explanation techniques can depend on the underlying model, the dataset, measure of similarity, which data point is explained, and what pre-processing technique that has been employed. Overall, LPI was observed to often provide more accurate explanations compared to both SHAP and LIME for both logistic regression and naive Bayes models when using Spearman's rank correlation whereas LIME provided the most accurate explanations when using Euclidean and Cosine similarity.

Based on our empirical experiments, our proposed measure of explanation accuracy is complementary to the functionally-grounded robustness measures. For example, LIME provides relatively low robustness values when explaining Logistic Regression and Naive Bayes models while outperforming LPI and SHAP with regards to explanation accuracy when the similarity metric of choice is Cosine or Euclidean similarity. \textit{Based on this, we argue that the notions of explanation robustness and accuracy are orthogonal concepts and should be studied and analyzed separately.}

The findings in this study point to several directions for future research. One major open question is how LPI is able to frequently outperform LIME and SHAP despite not even attempting to decompose the predictions into additive terms. Other open questions concern the interplay between the pre-processing technique, underlying model and explanation technique, e.g., how come the choice of pre-processing technique have an effect of the relative performance of the explanation techniques, and why only for one model class?

Another important direction for future research is to extend the proposed evaluation framework to other model classes, e.g., tree models, and explanation types, e.g., rules, as produced by Anchors \citep{ribeiro2018anchors}. One major challenge here is to derive the true feature importance scores in cases where intrinsic additive structures are not as easily derivable as they are for logistic regression and naive Bayes. 

\begin{acks}
To Robert, for the bagels and explaining CMYK and color spaces.
\end{acks}

\bibliographystyle{ACM-Reference-Format}
\bibliography{main}


\begin{thebibliography}{23}

%%% the Web address.

\ifx \showCODEN    \undefined \def \showCODEN     #1{\unskip}     \fi
\ifx \showDOI      \undefined \def \showDOI       #1{#1}\fi
\ifx \showISBNx    \undefined \def \showISBNx     #1{\unskip}     \fi
\ifx \showISBNxiii \undefined \def \showISBNxiii  #1{\unskip}     \fi
\ifx \showISSN     \undefined \def \showISSN      #1{\unskip}     \fi
\ifx \showLCCN     \undefined \def \showLCCN      #1{\unskip}     \fi
\ifx \shownote     \undefined \def \shownote      #1{#1}          \fi
\ifx \showarticletitle \undefined \def \showarticletitle #1{#1}   \fi
\ifx \showURL      \undefined \def \showURL       {\relax}        \fi
% invisible to TeX
\providecommand\bibfield[2]{#2}
\providecommand\bibinfo[2]{#2}
\providecommand\natexlab[1]{#1}
\providecommand\showeprint[2][]{arXiv:#2}

\bibitem[Alvarez-Melis and Jaakkola(2018)]%
        {alvarez2018robustness}
\bibfield{author}{\bibinfo{person}{David Alvarez-Melis} {and}
  \bibinfo{person}{Tommi~S Jaakkola}.} \bibinfo{year}{2018}\natexlab{}.
\newblock \showarticletitle{On the robustness of interpretability methods}.
\newblock \bibinfo{journal}{\emph{ICML Workshop on Human Interpretability in
  Machine Learning}} (\bibinfo{year}{2018}).
\newblock


\bibitem[Bishop(2006)]%
        {bishop2006pattern}
\bibfield{author}{\bibinfo{person}{Christopher~M Bishop}.}
  \bibinfo{year}{2006}\natexlab{}.
\newblock \showarticletitle{Pattern recognition}.
\newblock \bibinfo{journal}{\emph{Machine learning}} \bibinfo{volume}{128},
  \bibinfo{number}{9} (\bibinfo{year}{2006}).
\newblock


\bibitem[Camburu et~al\mbox{.}(2019)]%
        {camburu2019can}
\bibfield{author}{\bibinfo{person}{Oana-Maria Camburu},
  \bibinfo{person}{Eleonora Giunchiglia}, \bibinfo{person}{Jakob Foerster},
  \bibinfo{person}{Thomas Lukasiewicz}, {and} \bibinfo{person}{Phil Blunsom}.}
  \bibinfo{year}{2019}\natexlab{}.
\newblock \showarticletitle{Can I trust the explainer? Verifying post-hoc
  explanatory methods}.
\newblock \bibinfo{journal}{\emph{NeurIPS Workshop on Safety and Robustness in
  Machine Learning}} (\bibinfo{year}{2019}).
\newblock


\bibitem[Casalicchio et~al\mbox{.}(2018)]%
        {casalicchio2018visualizing}
\bibfield{author}{\bibinfo{person}{Giuseppe Casalicchio},
  \bibinfo{person}{Christoph Molnar}, {and} \bibinfo{person}{Bernd Bischl}.}
  \bibinfo{year}{2018}\natexlab{}.
\newblock \showarticletitle{Visualizing the feature importance for black box
  models}. In \bibinfo{booktitle}{\emph{Joint European Conference on Machine
  Learning and Knowledge Discovery in Databases}}. Springer,
  \bibinfo{pages}{655--670}.
\newblock


\bibitem[Chen et~al\mbox{.}(2018)]%
        {chen2018learning}
\bibfield{author}{\bibinfo{person}{Jianbo Chen}, \bibinfo{person}{Le Song},
  \bibinfo{person}{Martin Wainwright}, {and} \bibinfo{person}{Michael Jordan}.}
  \bibinfo{year}{2018}\natexlab{}.
\newblock \showarticletitle{Learning to explain: An information-theoretic
  perspective on model interpretation}. In
  \bibinfo{booktitle}{\emph{International Conference on Machine Learning}}.
  PMLR, \bibinfo{pages}{883--892}.
\newblock


\bibitem[Craven and Shavlik(1995)]%
        {craven1995extracting}
\bibfield{author}{\bibinfo{person}{Mark Craven} {and} \bibinfo{person}{Jude
  Shavlik}.} \bibinfo{year}{1995}\natexlab{}.
\newblock \showarticletitle{Extracting tree-structured representations of
  trained networks}.
\newblock \bibinfo{journal}{\emph{Advances in neural information processing
  systems}}  \bibinfo{volume}{8} (\bibinfo{year}{1995}),
  \bibinfo{pages}{24--30}.
\newblock


\bibitem[Doshi-Velez and Kim(2017)]%
        {doshi2017towards}
\bibfield{author}{\bibinfo{person}{Finale Doshi-Velez} {and}
  \bibinfo{person}{Been Kim}.} \bibinfo{year}{2017}\natexlab{}.
\newblock \showarticletitle{Towards a rigorous science of interpretable machine
  learning}.
\newblock \bibinfo{journal}{\emph{arXiv preprint arXiv:1702.08608}}
  (\bibinfo{year}{2017}).
\newblock


\bibitem[Fong and Vedaldi(2017)]%
        {fong2017interpretable}
\bibfield{author}{\bibinfo{person}{Ruth~C Fong} {and} \bibinfo{person}{Andrea
  Vedaldi}.} \bibinfo{year}{2017}\natexlab{}.
\newblock \showarticletitle{Interpretable explanations of black boxes by
  meaningful perturbation}. In \bibinfo{booktitle}{\emph{Proceedings of the
  IEEE international conference on computer vision}}.
  \bibinfo{pages}{3429--3437}.
\newblock


\bibitem[Freitas(2014)]%
        {freitas2014comprehensible}
\bibfield{author}{\bibinfo{person}{Alex~A Freitas}.}
  \bibinfo{year}{2014}\natexlab{}.
\newblock \showarticletitle{Comprehensible classification models: a position
  paper}.
\newblock \bibinfo{journal}{\emph{ACM SIGKDD explorations newsletter}}
  \bibinfo{volume}{15}, \bibinfo{number}{1} (\bibinfo{year}{2014}),
  \bibinfo{pages}{1--10}.
\newblock


\bibitem[Ghorbani et~al\mbox{.}(2019)]%
        {ghorbani2019interpretation}
\bibfield{author}{\bibinfo{person}{Amirata Ghorbani}, \bibinfo{person}{Abubakar
  Abid}, {and} \bibinfo{person}{James Zou}.} \bibinfo{year}{2019}\natexlab{}.
\newblock \showarticletitle{Interpretation of neural networks is fragile}. In
  \bibinfo{booktitle}{\emph{Proceedings of the AAAI Conference on Artificial
  Intelligence}}, Vol.~\bibinfo{volume}{33}. \bibinfo{pages}{3681--3688}.
\newblock


\bibitem[Hooker et~al\mbox{.}(2019)]%
        {hooker2018benchmark}
\bibfield{author}{\bibinfo{person}{Sara Hooker}, \bibinfo{person}{Dumitru
  Erhan}, \bibinfo{person}{Pieter-Jan Kindermans}, {and} \bibinfo{person}{Been
  Kim}.} \bibinfo{year}{2019}\natexlab{}.
\newblock \showarticletitle{A benchmark for interpretability methods in deep
  neural networks}.
\newblock \bibinfo{journal}{\emph{Advances in Neural Information Processing
  Systems 32 (NeurIPS)}} (\bibinfo{year}{2019}).
\newblock


\bibitem[Hsieh et~al\mbox{.}(2021)]%
        {hsieh2021evaluations}
\bibfield{author}{\bibinfo{person}{Cheng-Yu Hsieh}, \bibinfo{person}{Chih-Kuan
  Yeh}, \bibinfo{person}{Xuanqing Liu}, \bibinfo{person}{Pradeep Ravikumar},
  \bibinfo{person}{Seungyeon Kim}, \bibinfo{person}{Sanjiv Kumar}, {and}
  \bibinfo{person}{Cho-Jui Hsieh}.} \bibinfo{year}{2021}\natexlab{}.
\newblock \showarticletitle{Evaluations and Methods for Explanation through
  Robustness Analysis}.
\newblock \bibinfo{journal}{\emph{9th International Conference on Learning
  Representations, ICLR 2021}} (\bibinfo{year}{2021}).
\newblock


\bibitem[Lakkaraju et~al\mbox{.}(2020)]%
        {lakkaraju2020robust}
\bibfield{author}{\bibinfo{person}{Himabindu Lakkaraju}, \bibinfo{person}{Nino
  Arsov}, {and} \bibinfo{person}{Osbert Bastani}.}
  \bibinfo{year}{2020}\natexlab{}.
\newblock \showarticletitle{Robust and stable black box explanations}. In
  \bibinfo{booktitle}{\emph{International Conference on Machine Learning}}.
  PMLR, \bibinfo{pages}{5628--5638}.
\newblock


\bibitem[Lipton et~al\mbox{.}(2018)]%
        {lipton2018detecting}
\bibfield{author}{\bibinfo{person}{Zachary Lipton}, \bibinfo{person}{Yu-Xiang
  Wang}, {and} \bibinfo{person}{Alexander Smola}.}
  \bibinfo{year}{2018}\natexlab{}.
\newblock \showarticletitle{Detecting and correcting for label shift with black
  box predictors}. In \bibinfo{booktitle}{\emph{International conference on
  machine learning}}. PMLR, \bibinfo{pages}{3122--3130}.
\newblock


\bibitem[Lundberg and Lee(2017)]%
        {lundberg2017unified}
\bibfield{author}{\bibinfo{person}{Scott Lundberg} {and} \bibinfo{person}{Su-In
  Lee}.} \bibinfo{year}{2017}\natexlab{}.
\newblock \showarticletitle{A unified approach to interpreting model
  predictions}.
\newblock \bibinfo{journal}{\emph{Advances in Neural Information Processing
  Systems 30 (NeruIPS)}} (\bibinfo{year}{2017}).
\newblock


\bibitem[Montavon et~al\mbox{.}(2018)]%
        {montavon2018methods}
\bibfield{author}{\bibinfo{person}{Gr{\'e}goire Montavon},
  \bibinfo{person}{Wojciech Samek}, {and} \bibinfo{person}{Klaus-Robert
  M{\"u}ller}.} \bibinfo{year}{2018}\natexlab{}.
\newblock \showarticletitle{Methods for interpreting and understanding deep
  neural networks}.
\newblock \bibinfo{journal}{\emph{Digital Signal Processing}}
  \bibinfo{volume}{73} (\bibinfo{year}{2018}), \bibinfo{pages}{1--15}.
\newblock


\bibitem[Rahnama and Bostr{\"o}m(2019)]%
        {rahnama2019study}
\bibfield{author}{\bibinfo{person}{Amir Hossein~Akhavan Rahnama} {and}
  \bibinfo{person}{Henrik Bostr{\"o}m}.} \bibinfo{year}{2019}\natexlab{}.
\newblock \showarticletitle{A study of data and label shift in the LIME
  framework}.
\newblock \bibinfo{journal}{\emph{Neurip 2019 Workshop on Human-Centric Machine
  Learning}} (\bibinfo{year}{2019}).
\newblock


\bibitem[Ribeiro et~al\mbox{.}(2016a)]%
        {ribeiro2016should}
\bibfield{author}{\bibinfo{person}{Marco~Tulio Ribeiro},
  \bibinfo{person}{Sameer Singh}, {and} \bibinfo{person}{Carlos Guestrin}.}
  \bibinfo{year}{2016}\natexlab{a}.
\newblock \showarticletitle{" Why should i trust you?" Explaining the
  predictions of any classifier}. In \bibinfo{booktitle}{\emph{Proceedings of
  the 22nd ACM SIGKDD international conference on knowledge discovery and data
  mining}}. \bibinfo{pages}{1135--1144}.
\newblock


\bibitem[Ribeiro et~al\mbox{.}(2016b)]%
        {ribeiro2016model}
\bibfield{author}{\bibinfo{person}{Marco~Tulio Ribeiro},
  \bibinfo{person}{Sameer Singh}, {and} \bibinfo{person}{Carlos Guestrin}.}
  \bibinfo{year}{2016}\natexlab{b}.
\newblock \showarticletitle{Model-agnostic interpretability of machine
  learning}.
\newblock \bibinfo{journal}{\emph{ICML Workshop on Human Interpretability in
  Machine}} (\bibinfo{year}{2016}).
\newblock


\bibitem[Ribeiro et~al\mbox{.}(2018)]%
        {ribeiro2018anchors}
\bibfield{author}{\bibinfo{person}{Marco~Tulio Ribeiro},
  \bibinfo{person}{Sameer Singh}, {and} \bibinfo{person}{Carlos Guestrin}.}
  \bibinfo{year}{2018}\natexlab{}.
\newblock \showarticletitle{Anchors: High-precision model-agnostic
  explanations}. In \bibinfo{booktitle}{\emph{Proceedings of the AAAI
  conference on artificial intelligence}}, Vol.~\bibinfo{volume}{32}.
\newblock


\bibitem[Samek et~al\mbox{.}(2016)]%
        {samek2016evaluating}
\bibfield{author}{\bibinfo{person}{Wojciech Samek}, \bibinfo{person}{Alexander
  Binder}, \bibinfo{person}{Gr{\'e}goire Montavon}, \bibinfo{person}{Sebastian
  Lapuschkin}, {and} \bibinfo{person}{Klaus-Robert M{\"u}ller}.}
  \bibinfo{year}{2016}\natexlab{}.
\newblock \showarticletitle{Evaluating the visualization of what a deep neural
  network has learned}.
\newblock \bibinfo{journal}{\emph{IEEE transactions on neural networks and
  learning systems}} \bibinfo{volume}{28}, \bibinfo{number}{11}
  (\bibinfo{year}{2016}), \bibinfo{pages}{2660--2673}.
\newblock


\bibitem[Yang and Kim(2019)]%
        {yang2019benchmarking}
\bibfield{author}{\bibinfo{person}{Mengjiao Yang} {and} \bibinfo{person}{Been
  Kim}.} \bibinfo{year}{2019}\natexlab{}.
\newblock \showarticletitle{Benchmarking attribution methods with relative
  feature importance}.
\newblock \bibinfo{journal}{\emph{Neurip 2019 Workshop on Human-Centric Machine
  Learning}} (\bibinfo{year}{2019}).
\newblock


\bibitem[Zeiler and Fergus(2014)]%
        {zeiler2014visualizing}
\bibfield{author}{\bibinfo{person}{Matthew~D Zeiler} {and} \bibinfo{person}{Rob
  Fergus}.} \bibinfo{year}{2014}\natexlab{}.
\newblock \showarticletitle{Visualizing and understanding convolutional
  networks}. In \bibinfo{booktitle}{\emph{European conference on computer
  vision}}. Springer, \bibinfo{pages}{818--833}.
\newblock


\end{thebibliography}

\appendix

\onecolumn
\section{Appendix}
\subsection{Rank Correlation}
Table \ref{table:robust} and \ref{table:minmax} show the average Spearman's rank correlation of explanations with true importance scores when robust and minmax preprocessing is used. 

\begin{table*}[ht]
    \caption{Average Spearman's rank correlation between the true importance scores and local explanations for LIME, SHAP and LPI explanations when robust pre-processing is applied.}
        \begin{tabular}{l|rrr|rrr}
            \toprule
             Model $\xrightarrow{}$&  \multicolumn{3}{c}{Logistic Regression} &  \multicolumn{3}{|c}{Na\"ive Bayes} \\ \hline
            Dataset   &   LIME &   SHAP & LPI &   LIME &   SHAP & LPI\\ \hline
             Adult              & -0.075 &  0.372 & -0.104 & 0.706 & 0.396 & 0.75  \\
         Attrition          &  0.329 &  0.413 &  0.215 & 0.232 & 0.297 & 0.297 \\
         Audit              &  0.486 &  0.971 &  0.579 & 0.466 & 0.364 & 0.522 \\
         Banking            &  0.257 &  0.601 &  0.006 & 0.486 & 0.088 & 0.847 \\
         Banknote           &  0.798 &  0.924 &  0.79  & 0.892 & 0.78  & 0.902 \\
         Breast Cancer      &  0.737 &  0.977 &  0.756 & 0.698 & 0.683 & 0.454 \\
         Churn              &  0.279 &  0.227 &  0.378 & 0.744 & 0.537 & 0.649 \\
         Donors             &  0.034 &  0.351 &  0.278 & 0.131 & 0.311 & 0.559 \\
         HR                 & -0.003 &  0.254 &  0.208 & 0.783 & 0.275 & 0.676 \\
         Haberman           &  0.364 &  0.961 &  0.292 & 0.844 & 0.032 & 0.87  \\
         Hattrick           &  0.369 &  0.808 &  0.498 & 0.594 & 0.565 & 0.446 \\
         Heart Disease      &  0.183 &  0.253 &  0.152 & 0.797 & 0.424 & 0.624 \\
         Insurance          &  0.522 & -0.153 &  0.519 & 0.59  & 0.199 & 0.642 \\
         Iris               &  0.512 &  0.736 &  0.424 & 0.816 & 0.892 & 0.78  \\
         Loan               & -0.008 &  0.533 &  0.051 & 0.272 & 0.463 & 0.463 \\
         Pima Indians       &  0.712 &  0.985 &  0.481 & 0.738 & 0.47  & 0.603 \\
         Seismic            &  0.195 &  0.801 &  0.107 & 0.786 & 0.202 & 0.726 \\
         Spambase           &  0.142 &  0.967 &  0.024 & 0.794 & 0.194 & 0.473 \\
         Thera              &  0.024 &  0.8   &  0.085 & 0.283 & 0.48  & 0.48  \\
         Titanic            &  0.447 &  0.574 &  0.328 & 0.259 & 0.367 & 0.79  \\
         
            \hline
              \midrule
             Average            &  0.315 &  0.618 &  0.303 & 0.596 & 0.401 & 0.628 \\
         Standard Deviation &  0.254 &  0.319 &  0.243 & 0.235 & 0.214 & 0.161 \\
         \bottomrule
        \end{tabular}
   \label{table:robust}
\end{table*}

\begin{table*}[htb]
\caption{Average Spearman's rank correlation between the true importance scores and local explanations for explanations when explaining Logistic Regression and na\"ive Bayes Models when min-max pre-processing is applied}
\begin{tabular}{l|rrr|rrr}
    \toprule
     Model $\xrightarrow{}$&  \multicolumn{3}{c}{Logistic Regression} &  \multicolumn{3}{|c}{Na\"ive Bayes} \\ \hline
    Dataset   &   LIME &   SHAP & LPI &   LIME &   SHAP & LPI\\ \hline
     Adult              & -0.014 &  0.302 & -0.018 & 0.702 & 0.399 & 0.749 \\
     Attrition          &  0.261 &  0.237 &  0.213 & 0.25  & 0.297 & 0.297 \\
     Audit              & -0.01  &  0.746 & -0.152 & 0.551 & 0.622 & 0.506 \\
     Banking            & -0.043 &  0.342 &  0.293 & 0.478 & 0.089 & 0.848 \\
     Banknote           &  0.708 &  0.708 &  0.684 & 0.894 & 0.786 & 0.902 \\
     Breast Cancer      &  0.513 &  0.665 &  0.64  & 0.702 & 0.684 & 0.454 \\
     Churn              &  0.153 &  0.067 &  0.241 & 0.886 & 0.529 & 0.629 \\
     Donors             & -0.019 &  0.176 &  0.143 & 0.122 & 0.303 & 0.552 \\
     HR                 &  0.083 &  0.283 &  0.341 & 0.786 & 0.276 & 0.675 \\
     Haberman           & -0.219 &  0.36  & -0.319 & 0.831 & 0.032 & 0.87  \\
     Hattrick           &  0.273 &  0.44  &  0.329 & 0.614 & 0.563 & 0.449 \\
     Heart Disease      &  0.107 &  0.17  &  0.031 & 0.802 & 0.423 & 0.623 \\
     Insurance          &  0.4   & -0.069 &  0.418 & 0.601 & 0.2   & 0.635 \\
     Iris               &  0.187 &  0.135 &  0.059 & 0.84  & 0.892 & 0.78  \\
     Loan               & -0.203 &  0.514 & -0.088 & 0.366 & 0.463 & 0.463 \\
     Pima Indians       &  0.51  &  0.554 &  0.372 & 0.755 & 0.466 & 0.622 \\
     Seismic            &  0.167 &  0.785 &  0.079 & 0.774 & 0.198 & 0.718 \\
     Spambase           & -0.141 &  0.883 & -0.265 & 0.795 & 0.193 & 0.468 \\
     Thera              & -0.212 &  0.519 & -0.09  & 0.255 & 0.48  & 0.48  \\
     Titanic            &  0.503 &  0.123 &  0.382 & 0.825 & 0.38  & 0.794 \\
    \hline
      \midrule
    Average            &  0.15  &  0.397 &  0.165 & 0.641 & 0.414 & 0.626 \\
    Standard Deviation &  0.264 &  0.26  &  0.269 & 0.227 & 0.22  & 0.162 \\
 \bottomrule
\end{tabular}
    
   \label{table:minmax}

\end{table*}

\subsection{Robustness Measures}
\label{sec:appendix:robust}
In this section, the details of measuring robustness for each dataset is presented. Table \ref{table:nr} includes the result of Local Lipschitz measures. As stated earlier, robust explanations  have relatively low Local Lipschitz values. Table \ref{table:rri} includes the result of $\textrm{Robustness}-S_r$ whereas Table \ref{table:rru} shows the result of $\textrm{Robustness}-\bar{S}_r$  across all datasets. As stated earlier, larger values of $\textrm{Robustness}-S_r$ and lower values  $\textrm{Robustness}-\bar{S}_r$ are desirable for a robust explanation.

\begin{table*}[htb]
\caption{Local Lipschitz Values for each explanation techniques when explaining Logistic Regression and Naive Bayes model across all datasets}
\begin{tabular}{l|rrr|rrr}
    \toprule
     Model $\xrightarrow{}$&  \multicolumn{3}{c}{Logistic Regression} &  \multicolumn{3}{|c}{Na\"ive Bayes} \\ \hline
    Dataset   &   LIME &   SHAP & LPI &   LIME &   SHAP & LPI\\ \hline
     Adult              & 1.248 & 0.121 & 0.138 & 1.487 & 0.139 & 0.157 \\
     Attrition          & 0.58  & 0.114 & 0.067 & 0.448 & 0.143 & 0.116 \\
     Audit              & 5.886 & 0.002 & 0.006 & 5.873 & 0     & 0     \\
     Banking            & 0.744 & 0.068 & 0.155 & 2.537 & 0.06  & 0.177 \\
     Banknote           & 1.789 & 0.366 & 0.467 & 0.958 & 0.005 & 0.002 \\
     Breast Cancer      & 0.576 & 0.169 & 0.176 & 0.55  & 0.181 & 0.227 \\
     Churn              & 0.808 & 0.086 & 0.092 & 1.612 & 0.109 & 0.121 \\
     Donors             & 2.856 & 0.013 & 0.023 & 3.389 & 2.537 & 2.525 \\
     HR                 & 5.65  & 0.003 & 0.005 & 6.498 & 0     & 0     \\
     Haberman           & 2.723 & 0.254 & 0.41  & 0.014 & 2.105 & 5.668 \\
     Hattrick           & 1.355 & 0.073 & 0.205 & 0.07  & 1.921 & 6.656 \\
     Heart Disease      & 1.444 & 0.297 & 0.253 & 1.482 & 0.25  & 0.364 \\
     Insurance          & 2.33  & 0.041 & 0.088 & 6.366 & 0.039 & 0.157 \\
     Iris               & 0.438 & 0.231 & 0.197 & 1.035 & 0.128 & 0.133 \\
     Loan               & 2.474 & 0.214 & 0.217 & 2.563 & 0.238 & 0.296 \\
     Pima Indians       & 1.175 & 0.871 & 0.789 & 0.886 & 0.343 & 0.374 \\
     Seismic            & 5.202 & 0.263 & 0.253 & 9.003 & 0.28  & 0.432 \\
     Spambase           & 2.549 & 0.245 & 0.267 & 3.349 & 0.314 & 0.452 \\
     Thera              & 0.115 & 0.009 & 0.015 & 0.615 & 0.029 & 0.068 \\
     Titanic            & 3.453 & 0.004 & 0.005 & 3.688 & 0     & 0     \\
    \hline
      \midrule
     Average            & 2.17  & 0.172 & 0.191 & 2.621 & 0.441 & 0.896 \\
     Standard Deviation & 1.69  & 0.194 & 0.187 & 2.458 & 0.748 & 1.839 \\
 \bottomrule
\end{tabular}
    
   \label{table:nr}
\end{table*}

\begin{table*}[htb]
\caption{$\textrm{Robustness}-S_r$ Values for each explanation techniques when explaining Logistic Regression and Naive Bayes model across all datasets}
\begin{tabular}{l|rrr|rrr}
    \toprule
        Model $\xrightarrow{}$&  \multicolumn{3}{c}{Logistic Regression} &  \multicolumn{3}{|c}{Na\"ive Bayes} \\ \hline
    Dataset   &   LIME &   SHAP & LPI &   LIME &   SHAP & LPI\\ \hline
   Adult              & 0.062 & 0.059 & 0.064 & 0.067 & 0.058 & 0.067 \\
     Attrition          & 0.029 & 0.033 & 0.079 & 0.037 & 0.041 & 0.049 \\
     Audit              & 0.016 & 0.015 & 0.01  & 0.012 & 0.017 & 0.017 \\
     Banking            & 0.027 & 0.018 & 0.039 & 0.033 & 0.02  & 0.028 \\
     Banknote           & 0.251 & 0.251 & 0.251 & 0.267 & 0.267 & 0.267 \\
     Breast Cancer      & 0.096 & 0.095 & 0.106 & 0.105 & 0.104 & 0.108 \\
     Churn              & 0.027 & 0.01  & 0.025 & 0.054 & 0.025 & 0.058 \\
     Donors             & 0.235 & 0.174 & 0.248 & 0.447 & 0.391 & 0.602 \\
     HR                 & 0.006 & 0.014 & 0.002 & 0.001 & 0.022 & 0.022 \\
     Haberman           & 0.292 & 0.059 & 0.289 & 0.006 & 0.513 & 1.7   \\
     Hattrick           & 0.122 & 0.076 & 0.186 & 0.08  & 0.094 & 1.31  \\
     Heart Disease      & 0.179 & 0.191 & 0.294 & 0.045 & 0.101 & 0.112 \\
     Insurance          & 0.029 & 0.015 & 0.033 & 0.028 & 0.029 & 0.019 \\
     Iris               & 0.071 & 0.071 & 0.071 & 0.06  & 0.06  & 0.06  \\
     Loan               & 0.15  & 0.121 & 0.15  & 0.185 & 0.153 & 0.174 \\
     Pima Indians       & 0.294 & 0.294 & 0.294 & 0.222 & 0.222 & 0.222 \\
     Seismic            & 0.076 & 0.043 & 0.088 & 0.048 & 0.025 & 0.055 \\
     Spambase           & 0.131 & 0.105 & 0.148 & 0.139 & 0.122 & 0.195 \\
     Thera              & 0.003 & 0.002 & 0.004 & 0.011 & 0.006 & 0.01  \\
     Titanic            & 0.06  & 0.069 & 0.062 & 0.064 & 0.07  & 0.07  \\
    \hline
      \midrule
     Average            & 0.108 & 0.086 & 0.122 & 0.096 & 0.117 & 0.257 \\
     Standard Deviation & 0.094 & 0.081 & 0.101 & 0.108 & 0.132 & 0.441 \\
 \bottomrule
\end{tabular}
   \label{table:rri}
\end{table*}

\begin{table*}[htb]
\caption{$\textrm{robustness}-\hat{S}_r$ Values for each explanation techniques when explaining Logistic Regression and Naive Bayes model across all datasets}
\begin{tabular}{l|rrr|rrr}
    \toprule
     Model $\xrightarrow{}$&  \multicolumn{3}{c}{Logistic Regression} &  \multicolumn{3}{|c}{Na\"ive Bayes} \\ \hline
    Dataset   &   LIME &   SHAP & LPI &   LIME &   SHAP & LPI\\ \hline
    Adult              & 0.023 & 0.055 & 0.028 & 0.029 & 0.043 & 0.029 \\
     Attrition          & 0.111 & 0.11  & 0.033 & 0.019 & 0.008 & 0.042 \\
     Audit              & 0     & 0.001 & 0     & 0     & 0     & 0     \\
     Banking            & 0.016 & 0.017 & 0.037 & 0.022 & 0.009 & 0.053 \\
     Banknote           & 0     & 0     & 0     & 0     & 0     & 0     \\
     Breast Cancer      & 0.018 & 0.018 & 0.034 & 0.035 & 0.02  & 0.052 \\
     Churn              & 0.034 & 0.031 & 0.043 & 0.04  & 0.03  & 0.057 \\
     Donors             & 0.003 & 0.097 & 0.002 & 0.011 & 0.039 & 0.002 \\
     HR                 & 0     & 0.001 & 0     & 0     & 0     & 0     \\
     Haberman           & 0.272 & 0.182 & 0.255 & 0.61  & 0.61  & 0.02  \\
     Hattrick           & 0.117 & 0.219 & 0.05  & 0.674 & 0.674 & 0.674 \\
     Heart Disease      & 0.009 & 0.039 & 0.017 & 0.062 & 0.06  & 0.04  \\
     Insurance          & 0.02  & 0.011 & 0.022 & 0.023 & 0.004 & 0.089 \\
     Iris               & 0     & 0     & 0     & 0     & 0     & 0     \\
     Loan               & 0.011 & 0.096 & 0.014 & 0.016 & 0.072 & 0.015 \\
     Pima Indians       & 0     & 0     & 0     & 0     & 0     & 0     \\
     Seismic            & 0.05  & 0.017 & 0.052 & 0.048 & 0.009 & 0.048 \\
     Spambase           & 0.082 & 0.073 & 0.059 & 0.076 & 0.051 & 0.089 \\
     Thera              & 0.002 & 0.003 & 0.004 & 0.009 & 0.003 & 0.015 \\
     Titanic            & 0.001 & 0.007 & 0.001 & 0     & 0     & 0     \\
    \hline
      \midrule
    Average            & 0.038 & 0.049 & 0.033 & 0.084 & 0.082 & 0.061 \\
    Standard Deviation & 0.064 & 0.062 & 0.055 & 0.188 & 0.188 & 0.143 \\
 \bottomrule
\end{tabular}
   \label{table:rru}
\end{table*}

\subsection{Accuracy}
\label{sec:appendix:accuracy}
The test accuracy of Logistic Regression and Naive Bayes models across all datasets is reported in Table \ref{table:accuracy}.

\begin{table*}[htb]
\caption{Test accuracy of Logistic Regression and Naive Bayes models using different preprocessing techniques across all datasets}
  \centering
\label{table:accuracy}
\begin{tabular}{lcc|cc|cc}
    \toprule
     Pre-processing &  \multicolumn{2}{c}{Standard} &  \multicolumn{2}{|c}{Robust} &  \multicolumn{2}{|c}{Minmax} \\ \hline
     dataset &   LR &   NB & LR &   NB &   LR & NB \\ \hline
     Attrition     & 1     & 1     & 1     & 1     & 1     & 1     \\
     Breast Cancer & 0.965 & 0.916 & 0.965 & 0.916 & 0.965 & 0.916 \\
     Pima Indians  & 0.802 & 0.766 & 0.776 & 0.766 & 0.802 & 0.766 \\
     Banknote      & 0.98  & 0.854 & 0.977 & 0.854 & 0.98  & 0.854 \\
     Iris          & 1     & 1     & 1     & 1     & 1     & 1     \\
     Haberman      & 0.662 & 0.649 & 0.688 & 0.649 & 0.662 & 0.649 \\
     Spambase      & 0.921 & 0.808 & 0.886 & 0.808 & 0.921 & 0.809 \\
     Adult         & 0.841 & 0.805 & 0.835 & 0.805 & 0.841 & 0.801 \\
     Heart Disease & 0.829 & 0.829 & 0.829 & 0.829 & 0.829 & 0.829 \\
     Churn         & 0.805 & 0.826 & 0.806 & 0.826 & 0.805 & 0.826 \\
     Hattrick      & 1     & 0.939 & 0.995 & 0.939 & 0.998 & 0.939 \\
     HR            & 0.77  & 0.749 & 0.77  & 0.749 & 0.77  & 0.749 \\
     Insurance     & 0.987 & 0.953 & 0.987 & 0.953 & 0.987 & 0.953 \\
     Audit         & 0.99  & 0.974 & 0.974 & 0.969 & 0.995 & 0.974 \\
     Loan          & 1     & 1     & 1     & 1     & 1     & 1     \\
     Donors        & 1     & 0.99  & 1     & 0.99  & 1     & 0.988 \\
     Seismic       & 0.946 & 0.837 & 0.947 & 0.837 & 0.944 & 0.837 \\
     Thera         & 1     & 1     & 1     & 1     & 1     & 1     \\
     Banking       & 0.906 & 0.879 & 0.905 & 0.879 & 0.907 & 0.879 \\
     Titanic       & 0.794 & 0.78  & 0.794 & 0.78  & 0.794 & 0.78  \\
 \bottomrule
\end{tabular}
\end{table*}

\subsection{Na\"ive Bayes example}
In this section, we show an example of how LOR scores are extracted for a Naive Bayes model. Let us train a Gaussian Na\"ve Bayes model on the following data and label matrix: 

\begin{equation*}
    X = \begin{pmatrix}
    -1 & -1 \\
    -2 & -1 \\
    -3 & -2 \\
    1 & 1 \\
    2 & 1 \\
    3 & 2 
    \end{pmatrix}
    , 
    Y = \begin{pmatrix}
    0 \\
    0 \\
    0 \\
    1 \\
    1 \\
    1  
    \end{pmatrix}
\end{equation*}

\begin{figure}
 \centering
  \includegraphics[width=0.49\textwidth]{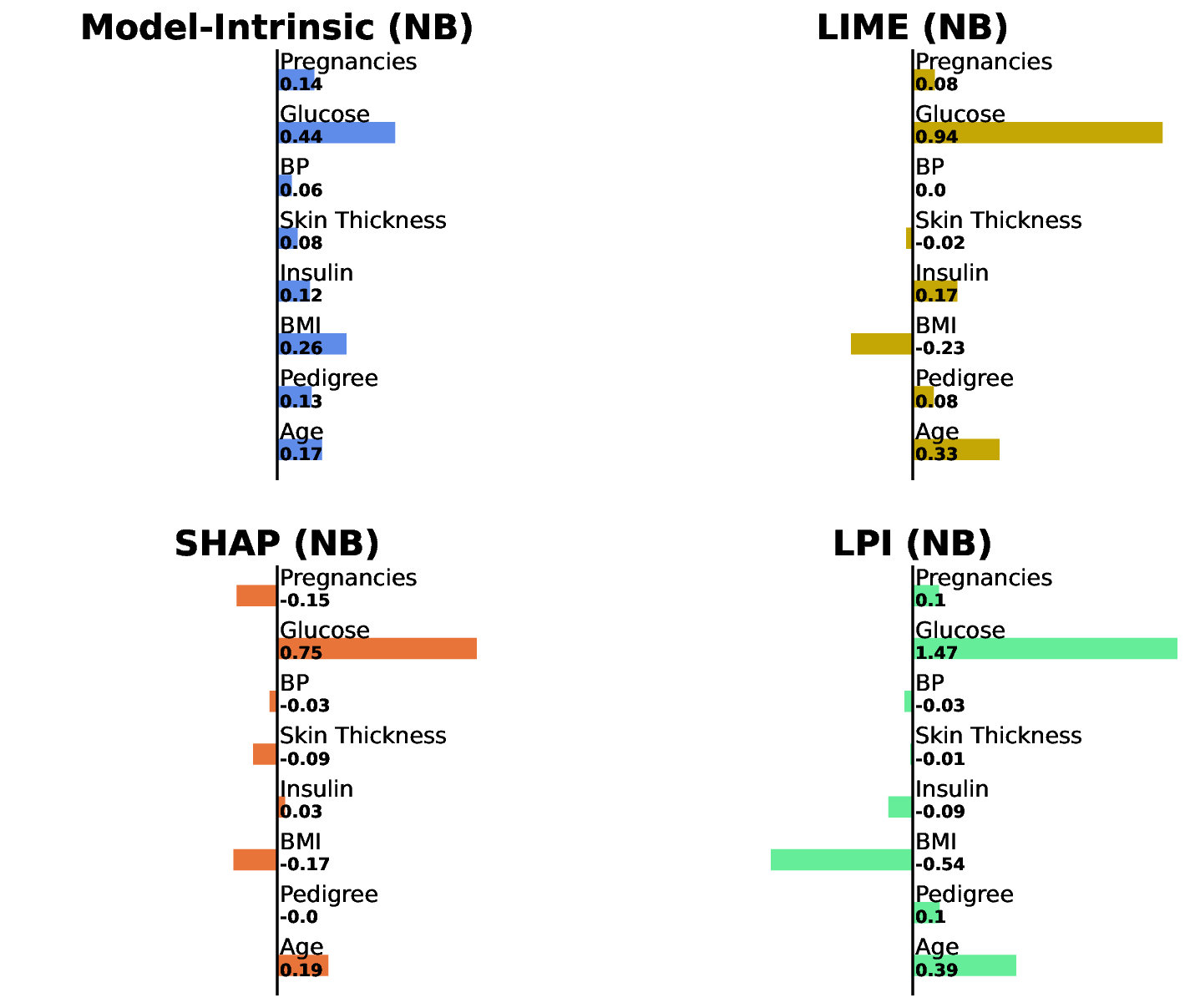}
 \caption{The feature importance scores of aLOR as well as LIME, SHAP and LPI explanations for a single instance from the Pima Indians data set when explaining a Naive Bayes prediction.}
  \label{fig:nb_exp}
 \end{figure}

The parameters of the Gaussian distribution for feature 1 and 2 for class 0 are $\mathcal{N}(-2.0, 0.66)$ and $\mathcal{N}(-1.33, 0.22)$. Similarly, parameters of the Gaussian distribution for feature 1 and 2 for class 1 are: $\mathcal{N}(2.0, 0.66)$ $\mathcal{N}(1.33, 0.22)$. For $x_n$ with $x_n^1 = -2$ and $x_n^2 = -1$, the model predicts $P(y_n=1|x_n) = 1$. Let $c=0$, therefore

\begin{align*}
    \mathcal{N}(x_n^0|\mu_{c}^0, \sigma_{c}^0) = 0.488\\
    \mathcal{N}(x_n^1|\mu_{c}^1, \sigma_{c}^1) =   3.002 \textbf{e}^{-6} \\
    \mathcal{N}(x_n^0|\mu_{\neg c}^0, \sigma_{\neg c}^0) = 0.65 \\
    \mathcal{N}(x_n^1|\mu_{\neg c}^1, \sigma_{\neg c}^1)= 4.04 \textbf{e}^{-6}
\end{align*}

based on this, 

\begin{align*}
    \textrm{log}(\frac{1}{3.7 e^{-11}}) = \textrm{log}\frac{0.488}{3.002 \textbf{e}^{-6}} \\
    +  \textrm{log}\frac{0.65}{4.04 \textbf{e}^{-6}} \\
    23.99 = 11.99 + 11.99
\end{align*}

where $const =log(1)$. While the first feature has an average of -2 for class 0 in the global Gaussian distribution parameters, the contribution of this feature to the LOR of Na\"ve Bayes model for $x_n$ is largely positive, i.e. 11.99. From these two examples, we can see that beside providing true local importance scores, our proposed method can help to quantify the differences between the global and local importance scores in Naive Bayes. Figure \ref{fig:nb_exp} shows another example where the extracted LOR scores are compared against different explanations for a test instance in Pima Indians dataset.

\subsection{Datasets}
Descriptions and access links to all datasets used in this study is available in Table \ref{table:datasets}.

\begin{table*}[htb]
\scriptsize
  \caption{List of datasets used in this study}
  \label{table:datasets}
  \centering
    \begin{tabular}{lr}

    \toprule
          dataset     &   URL  \\
        \midrule
         Adult         &      \url{https://archive.ics.uci.edu/ml/datasets/adult}\\
         Attrition     &      \url{https://www.kaggle.com/philschmidt/employee-attrition-eda} \\
         Audit         &      \url{https://www.kaggle.com/sid321axn/audit-data}\\
         Banking       &      \url{https://www.kaggle.com/rashmiranu/banking-dataset-classification}\\
         Banknote      &     \url{https://archive.ics.uci.edu/ml/datasets/banknote+authentication}  \\
         Breast Cancer &      \url{https://archive.ics.uci.edu/ml/datasets/breast+cancer+wisconsin+(diagnostic)}\\
         Churn         &      \url{https://www.kaggle.com/surendharanp/personal-loan}\\
         Donors        &      \url{https://www.kaggle.com/momohmustapha/donorsprediction}\\
         Pima Indians  &      \url{https://www.kaggle.com/uciml/pima-indians-diabetes-database} \\
         Haberman      &      \url{https://archive.ics.uci.edu/ml/datasets/haberman's+survival}\\
         Hattrick      &      \url{https://www.kaggle.com/juandelacalle/hattirckorg-matches-dataset}\\
         Heart Disease &      \url{https://archive.ics.uci.edu/ml/datasets/heart+disease}\\
         HR            &      \url{https://www.kaggle.com/sid321axn/audit-data}\\
         Insurance     &      \url{https://www.kaggle.com/mhdzahier/travel-insurance}\\
         Iris          &      \url{https://archive.ics.uci.edu/ml/datasets/iris}\\
         Loan          &      \url{https://www.kaggle.com/teertha/personal-loan-modeling}\\
         Seismic       &      \url{https://archive.ics.uci.edu/ml/datasets/seismic-bumps}\\
         Spambase      &      \url{https://archive.ics.uci.edu/ml/datasets/spambase}\\
         Thera         &      \url{https://www.kaggle.com/surendharanp/personal-loan}\\
         Titanic       &      \url{https://www.kaggle.com/c/titanic}\\
    \bottomrule
    \end{tabular}
\end{table*}

\subsection{Preprocessing techniques}
\label{appendix:prproc_def}
We have included two additional pre-processing techniques including min-max scaling and inter-quartile pre-processing. In inter-quartile pre-processing, median is removed and data is scaled according to the quantile range between the first quartile and the third quartile. For a given random variable $x$, min-max scaling produces $x'$ as follows:

\begin{equation*}
    x' = \frac{x - \textrm{min}(x)}{\textrm{max}(x) - \textrm{min}(x)}
\end{equation*}

\end{document}